\documentclass{article}

\usepackage{PRIMEarxiv}

\usepackage[utf8]{inputenc} 
\usepackage[T1]{fontenc}    
\usepackage{hyperref}       
\usepackage{url}            
\usepackage{booktabs}       
\usepackage{amsfonts}       
\usepackage{nicefrac}       
\usepackage{microtype}      
\usepackage{lipsum}
\usepackage{fancyhdr}       
\usepackage{graphicx}       
\graphicspath{{media/}}     

\usepackage{natbib}
 \bibpunct[, ]{(}{)}{,}{a}{}{,}%

\usepackage[parfill]{parskip}
\usepackage{multirow}
\usepackage[ruled, vlined, linesnumbered]{algorithm2e}
\usepackage{romannum}
\usepackage{placeins}
\usepackage{csquotes}
\usepackage{paralist}
\usepackage{xcolor}
\usepackage[onehalfspacing]{setspace}
\usepackage{appendix}
\usepackage{subcaption}

\RequirePackage{amsmath,amssymb,ifthen,array,theorem}
\def\argmax{\mathop{\rm arg\,max}}%
\def\argmin{\mathop{\rm arg\,min}}%

\newtheorem{proposition}{Proposition}[section]
\newtheorem{corollary}{Corollary}[section]

\newenvironment{proof}{\paragraph{Proof:}}{\hfill$\square$}

\AtBeginDocument{\pagenumbering{arabic}}

\definecolor{safeblue}{RGB}{0,0,0}

\title{Learning State-Dependent Policy Parametrizations\\ for Dynamic Technician Routing with Rework}

\author{
  Jonas Stein, Florentin D Hildebrandt, Marlin W Ulmer \\
  Chair of Management Science \\
  Otto-von-Guericke-Universität Magdeburg \\
  Magdeburg\\
  \texttt{\{jonas.stein, florentin.hildebrandt, marlin.ulmer\}@ovgu.de} \\
   \And
    Barrett W Thomas \\
  Tippie College of Business  \\
  University of Iowa \\
  Iowa\\
  \texttt{barrett-thomas@uiowa.edu} \\}

\begin{document}
\maketitle

\begin{abstract}
Home repair and installation services require technicians to visit customers and resolve tasks of different complexity. Technicians often have heterogeneous skills and working experiences. The geographical spread of customers makes achieving only perfect matches between technician skills and task requirements impractical. Additionally, technicians are regularly absent due to sickness. With non-perfect assignments regarding task requirement and technician skill, some tasks may remain unresolved and require a revisit and rework. Companies seek to minimize customer inconvenience due to delay. We model the problem as a sequential decision process where, over a number of service days, customers request service while heterogeneously skilled technicians are routed to serve customers in the system. Each day, our policy iteratively builds tours by adding \enquote{important} customers. The importance bases on analytical considerations and is measured by respecting routing efficiency, urgency of service, and risk of rework in an integrated fashion. We propose a state-dependent balance of these factors via reinforcement learning. A comprehensive study shows that taking a few non-perfect assignments can be quite beneficial for the overall service quality. We further demonstrate the value provided by a state-dependent parametrization.
\end{abstract}


\section{Introduction}\label{sec: introduction}

One of the authors recently bought a new kitchen. After installation, the lighting did not work correctly and the after-sales department of the kitchen company sent a technician to fix it. It did not work, and a few days later, another technician tried their luck, also unsuccessfully. Finally, the company sent one of their experts who resolved the issue within minutes. All this led to significant labor cost for the company and frustration for the author. The issue of unresolved repair or installation services and repeated technician visits for rework is not unique to kitchen installation. It is also common in other services with complex technician tasks such as home appliance repair, the repair of home electrics, or the installation of cable \citep{vanMoeseke2022}. One reason can be a missing spare part \citep{Pham2024}. This issue can be resolved with remote diagnostic tools that can communicate the reason for failure quite accurately and the respective spare parts can be loaded to the technician's truck \citep{Rippe2023}. However, even if the theoretical issue is known, there are a lot of practical circumstances involved. Here, another reason for services remaining unresolved is the individual skill level of the sent technicians \citep{Han2023}. Some \textit{expert} technicians have long term experience or additional training, but they are rare and expensive. In Germany's craft sector, the skilled labor shortage has led to an all-time high of more than 10,000 vacant positions for \textit{expert} technicians in 2022 \citep{Malin2023}. Simultaneously, the number of apprentices has decreased by more than 25$\%$ in the last 15 years, with a continuous downward trend \citep{Statista2024}. Thus, companies today, and even more so in the future, also employ \textit{regular} technicians who are able to fix \textit{easy} everyday problems but may not be able to resolve more \textit{advanced} issues with certainty. In the mentioned kitchen lighting case, due to a mismatch between the technician's skill and the task difficulty, the issue remained unresolved twice.

So, why not simply assign \textit{expert} technicians to \textit{advanced} tasks and \textit{regular} technicians to everyday issues to avoid rework? One reason is that technicians, as all employees, call in sick regularly \citep{Khalfay2017}. In Germany, employees working in the maintenance and repair sector called in sick for almost five weeks in 2022 \citep{IWD2024}. Thus, even when a perfect assignment of technicians and skills may be theoretically possible, technician absences may thwart the companies plans. Still, even at days were all technicians are available, perfect assignments are often not possible due the large geographical spread of customers. The aforementioned kitchen company operates in a large service area with travel distances from the headquarter to customers of often more than 100 kilometers. This geographical spread makes such an exclusive, skill-oriented assignment of technicians to tasks quite inefficient. Only few tasks can be performed per day, leaving many customers waiting for service. The alternative, disregarding technicians' skill levels to determine the most efficient route, could allow for many visits. However, it may also lead to significant rework, leaving many customers waiting again for extended periods until successful service completion. All this while every day, more customers call for fast and reliable technician services. In practice, such issues often not only lead to unhappy customers, but also to companies being unable to accommodate new customers at some point due to the accumulated workload \citep{Stock2024}.

In this work, we seek to balance skill-oriented assignments and effective routing while avoiding exceptionally long waiting for individual customers. Our goal is to minimize the average inconvenience experienced by customers for late service completions. We present a policy that iteratively builds a tour with respect to routing efficiency, assignment quality, and customer inconvenience due to delay. The factors are balanced in a score function motivated by analytical considerations. Since the importance of the factors may be state-dependent, we learn a state-dependent parametrization of this function via reinforcement learning (RL). In a comprehensive computational study, we derive the following managerial insights:
\begin{compactitem}
  \setlength{\parskip}{0pt}
  \setlength{\parsep}{0pt}
    \item Ignoring heterogeneity in tasks and workforce skills leads very poor customer experience.
    \item Avoiding inconveniences by any means demands considerable resources and proves unsustainable in the long term due to a congested system. 
    \item Same is true for avoiding any non-perfect, \textit{risky}, assignments. When done right, carefully relaxing the perfect, \textit{safe}, assignment constraint for a few customers (about 7\%) benefits company and customers.
    \item Splitting the workforce into \textit{regular} and \textit{expert} technicians, and categorizing tasks as \textit{easy} or \textit{advanced}, works surprisingly well when each group is handled individually. It is effective when the ratios of workforce types to task requirements are similar (for e.g., 50\% \textit{regular} technicians, 50\% \textit{easy} tasks).
    \item Effective decision-making requires a careful and state-dependent balance between routing efficiency, skill-oriented assignments, and avoidance of inconvenience. Individually focusing on one of them leads to poor decisions and many dissatisfied customers.
    \item A state-dependent parametrization leads to an improvement of almost $8\%$ in our experiments compared to a static parametrization. In states with many delayed customers, the focus should shift toward efficient routing. Conversely, in states with few urgent customers, avoiding inconvenience should be prioritized.
    \item All customers experience a similar level of service, regardless if they are located conveniently close to the company's location or far away in remote areas.
    \item Fewer \textit{expert} technicians are needed to achieve similar performances compared to benchmark policies.
    \item The success rate of technicians is more than $95\%$, likely increasing job satisfaction for technicians.
    \item Our policy outperforms benchmark policies regardless of the technician absence rate. However, the absence likelihood still has a significant impact on service quality.
\end{compactitem}

\noindent Our work makes the following problem-oriented and methodological contributions:

\paragraph{Problem.} Our problem is among the first in service routing that explicitly models driver absences and considers rework uncertainty due to heterogeneous workforce skills. We present an unambiguous mathematical model and an effective policy tailored to the problem characteristics. We provide practical insights with respect to important problem dimensions. 

\paragraph{Methodology.} Our proposed method integrates intuitive decision making via a score-based assignment strategy and resolute dynamic optimization via RL. While the general concept is introduced in \cite{Hildebrandt2023}, we are the first to present a dynamic policy tuned state-dependently via RL and to investigate its detailed functionality for a dynamic routing problem of practical complexity. We do not only show that RL can be used for state-dependent parametrization, but we also illustrate and analyze the specific algorithmic augmentations required to make it work.

Since problem and methodology are novel, we make specific suggestions for promising follow-up research. The remainder of this work is structured as follows. In Section~\ref{sec: literature}, we provide the literature overview related to our problem. Section~\ref{sec: problem} introduces our problem and defines the model, followed by Section~\ref{sec: methodology} with the solution approach. We present our computational evaluation in Section~\ref{sec: evaluation} and provide a conclusion and motivation for future research areas in the last Section~\ref{sec: conclusion}. Appendix~\ref{sec: appendix} provides additional details.

\section{Related Literature}\label{sec: literature}

In our literature review, we present papers that closely align with our work. In Appendix \ref{app: literature}, we summarize the large body of work on deterministic and multi-period service routing with heterogeneous workforce, and broader routing problems that incorporate repeated customer visits.

As our work, \cite{Chen2016} consider the routing of technicians in a stochastic and dynamic context. On a daily basis, a set of technician routes is scheduled to provide on-site services to newly requesting customers with the goal of minimizing completion time. Over time, technicians are able to improve their qualifications through learning, resulting in decreased service and completion times. By tuning a cost function approximation to handle anticipation more directly, the authors demonstrate that effective scheduling decisions should consider the individual learning rates of the technicians. As in our work, the authors assign heterogeneous technicians to tasks of varying difficulty. However, similar to the existing body of literature, it is assumed that any technician can resolve any task with certainty, though with varying service times and no rework.

\cite{Pham2024} introduce an optimization problem involving spare parts planning and technician scheduling for companies providing repair services for large household appliances. Over a sequence of days, a single technician visits dynamically requesting customers whose equipment has failed and needs repair. The spare parts required are stochastic since they are only revealed when a technician first arrives at the customer. Technicians load a variety of spare parts into their vans before departing from the depot. If a technician does not have all the required spare parts available, the repair remains unresolved and the customer needs to be revisited on a future day. The goal is to minimize the average daily costs which involve travel, holding, delay and repair costs. The authors develop anticipatory solution methods based on a value function approximation technique. The work shows similarities to ours since customer services can fail. However, the key differences are that we model both a fleet and heterogeneous technicians, and our method is designed to tackle the resulting assignment decisions.

\cite{Khorasanian2024} introduce a dynamic home care routing and scheduling problem. Over a rolling planning horizon, a single nurse is assigned to referrals with each requiring multiple visits for treatment. On a daily basis, the nurse performs subsequent visits to patients located within a defined service area. The daily number of new referrals as well as their required number of visits are unknown. New referrals can be rejected which comes with rejection costs. The goal is to minimize the overall cost of service. The authors propose a reinforcement learning policy to decide about rejection of customers. They show that an effective policy rejects (all) customers who are too far away from the depot. We employ a similar strategy in one of our benchmark policies (see \textit{MYEF} in Section~\ref{sec: benchmark}). It proves ineffective in our experiment since we cannot reject customers, and the inconvenience for customers who are located farther away becomes high if we decide against serving them early.

\cite{Nowak2024} focus on technician routing in the sharing economy, an environment where technicians operate autonomously from their own homes. Over multiple days, customers may require multiple services and moreover, task can require more than one day of work. Technicians need to have the required skill to complete a task, preventing any mismatched assignments and potential service failures. In our experiments, we employ a similar solution policy (see \textit{MYEX} in Section~\ref{sec: benchmark}) that emphasizes skill-oriented assignments, leading to promising results under certain conditions. The authors conduct a multi-objective approach considering routing, service and preparation costs. Additionally, as in our study, delayed services are determined and integrated as costs. Within a computational analysis evaluating different problem instances (e.g., fleet size, technician skills, task types), results show that a dispatched fleet of highly specialized technicians maximizes the served demand. 

\cite{Wolbeck2020} introduce a nurse rescheduling problem, taking into consideration schedule disruptions, e.g., spontaneous absences of nurses due to illness. Consequently, changes in shifts for nurses are required to fulfill the demand. The authors managed to implement a robust, general optimization model to establish a fair distribution of shift changes among nurses. Inspired by these ideas, we also incorporate workforce absences due to illness of technicians. Uncertain workforce availability is particularly common in crowdsourced deliveries, a sector where couriers are employed as gig workers \citep{Savelsbergh2022}. Dispatcher responsible for assigning customers and routing couriers faces uncertainties in the availability of workforce as couriers decide individually when to work.

In summary, while there is related work on dynamic routing of heterogeneous workforces, we are the first who consider risks of rework due to the lack of skill and technician absences in model and methodology. 

\section{Problem Definition}\label{sec: problem}

In this section, we present the problem description. We give a formal definition of the decision problem followed by an illustrative example. Afterwards, we model our problem as a sequential decision process following the framework by \cite{Powell2021}.

\subsection{Formal Problem Description}\label{sec: formal description}

Over multiple periods (e.g., days in one month), technicians serve customers within a defined service area. The set of technicians is fixed and their working hours per period are limited. Each technician has the same probability to be unavailable on a given day due to sickness. Technicians also possess different qualification levels. Some \textit{regular} technicians are less qualified due to limited practical experience or training. Other \textit{expert} technicians are highly qualified with extensive work experience. 

At the beginning of each period, customers request on-site technician services. Each request contains three elements: the location within the service area, the task difficulty, and the period the service is due (deadline). Related to the task difficulty, we define two levels. The first level comprises manageable \textit{easy} tasks that do not require advanced skills of technicians. Conversely, the second level comprises \textit{advanced} tasks that require higher demands on technicians. In our work, we assume that we know the task levels with certainty and that all tasks come with same service times. Besides the tasks, customers have deadlines, e.g., two days after the request was made, representing a threshold for delayed services. Customers are served on time if their requests are completed successfully before the deadline period, and vice versa, services are considered late when completed after the deadline. Starting from the deadline, customers experience increasing inconvenience provoked by extended waiting times and unreliable service promises.

Each period, the dispatcher determines which customers to serve, assigns customers to the available technicians, and plans their routes while considering the working time limitations. If assigned, customers with \textit{easy} tasks are served successfully regardless by the assigned technician. Customers with \textit{advanced} tasks are served with certainty if assigned to an \textit{expert} technician, but the tasks may remain unresolved and require rework with a known probability if assigned to a \textit{regular} technician. As a result, we have two types of assignments. \textit{Safe} assignments (\textit{advanced} task to \textit{expert} technician, \textit{easy} task to any technician) result in service completion whereas the successful outcome of \textit{risky} assignments (\textit{advanced} to \textit{regular}) is uncertain.

After each period, unassigned customers are postponed to the next period and customers with \textit{safe} assignments leave the system. For customers with \textit{risky} assignments, the success of the service realizes and the customers with unresolved services are also postponed to the next period. All postponed customers with violated deadlines experience inconvenience, measured in the time elapsed since the deadline. Routing itself does consume time and thus affects the ability to complete work within limited working resources. However, since technicians have fixed contracts, costs associated with routing are considered secondary and are omitted from the objective function. At the next period, new customer requests realize and are added to the postponed customers. To model the issue of companies not accommodating new customers anymore due to accumulated work, we stop considering new customer requests after a known period \citep{Stock2024}. Subsequent periods, which encompass the \textit{leftover} phase, are only used for completing all remaining requests. We note that for more effective policies, such a stop may occur later (or not at all) depending on the demand and given resources. However, we opted for this modeling choice to allow for a clear comparison between policies. The goal is to minimize the total inconvenience experienced by the customers. Inconvenience increases each day that a customer waits for service.

\subsection{Example}\label{sec: example}

\begin{figure}[!t]
    \centering
    \includegraphics[width = 1.0 \textwidth]{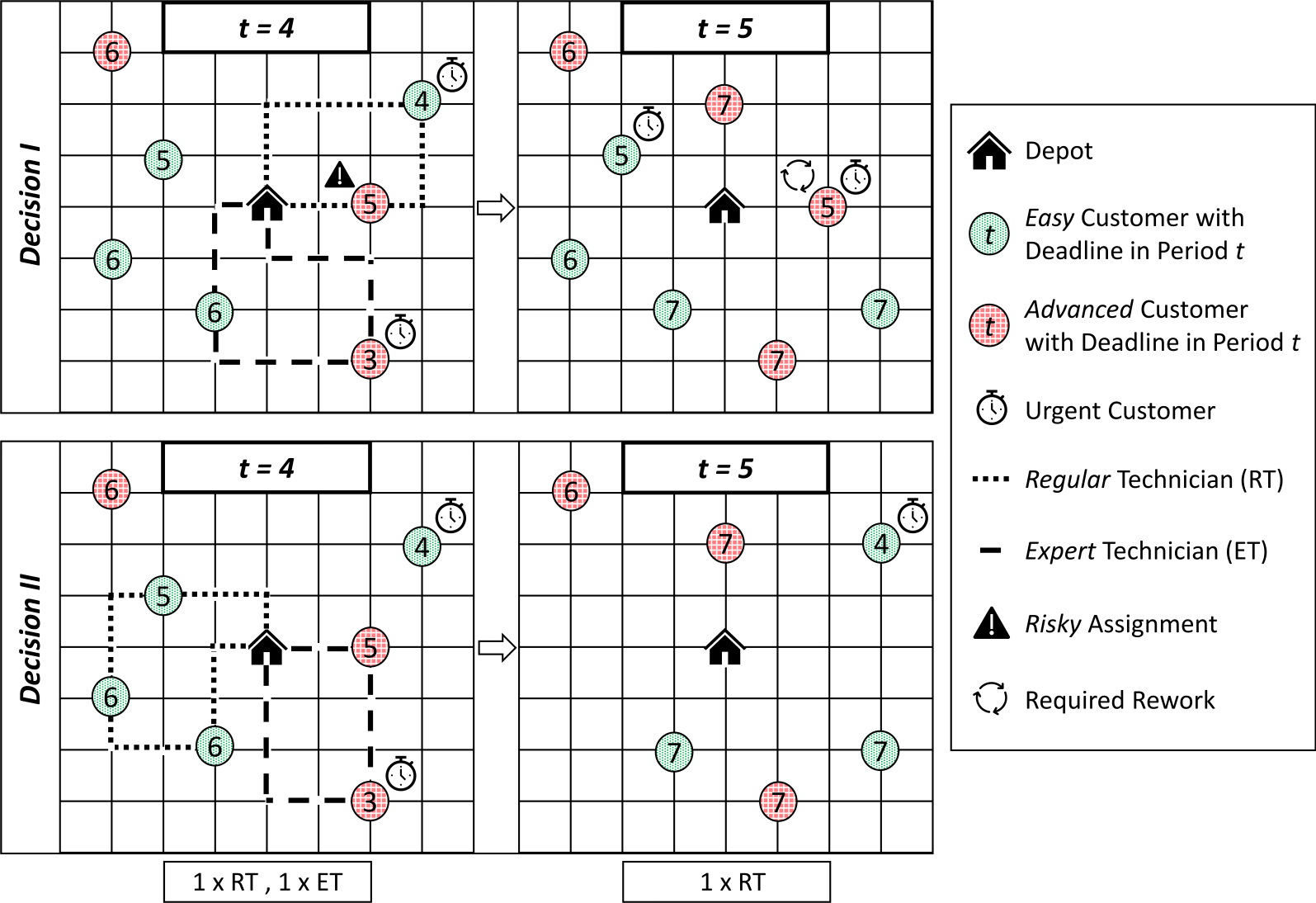}
    \caption{\centering Decisions and resulting next states; I (top) and II (bottom)}
    \label{fig: decisions}
\end{figure}

In this section, we illustrate the decision process with a small example for a decision state with two potential decisions and their outcomes (see Figure~\ref{fig: decisions}). For the purpose of presentation, we assume a Manhattan-style grid with travel times of 10 minutes per segment. The depot is located in the center. Service times are set to 30 minutes and technicians have a maximum working time of 210 minutes. We further assume, (all) two technicians are available in this period, one \textit{regular} and one \textit{expert}. The dotted and dashed lines show the tours for the \textit{regular} and \textit{expert} technician, respectively. The customer inconvenience grows by 10\% for each day a service is delayed.
The decision state is set in period $t=4$. At that period, seven customers are in the system illustrated by the circles. The (green) scatter plots in the circles represent \textit{easy} customers while the (red) dot matrix represents \textit{advanced} customers. The individual deadline periods are represented by the numbers in the circles. In this example, one customer's deadline has already been exceeded by one period, and another customer has a deadline in the current period. Both represent urgent customers, as leaving their requests incomplete would cause (additional) inconvenience. Two customers have their deadlines in the next period, and three customers have their deadlines in the period after next.

On the left side of Figure~\ref{fig: decisions}, we visualize two identical decision states with two different decisions. A decision involves assigning customers to technicians and determining their routing. In Decision I, top left of Figure~\ref{fig: decisions}, the \textit{regular} technician visits two customers in the northeast of the service area. The working time is 160 minutes (10 $\times$ 10 minutes + 2 $\times$ 30 minutes) and therefore feasible. The assigned \textit{advanced} customer is a \textit{risky} assignment which might result in rework. The \textit{expert} technician visits one \textit{easy} and one \textit{advanced} customer in the south with a working time of 180 minutes (12 $\times$ 10 + 2 $\times$ 30). Both assignments are \textit{safe}. The remaining three customers in the west are not visited in the period. The top right part of Figure~\ref{fig: decisions} shows a potential realization of uncertainty and the resulting next state in period $t=5$ following Decision I. A realization contains three parts: the outcomes of \textit{risky} assignments, new requests and the technician availability in the next period. In the example, we see that the \textit{risky} assignment remains unresolved and the corresponding customer located east to the depot requires rework. We further observe four new customers that request services, each with a deadline in period $t=7$. The \textit{expert} technician is absent in this state.

The alternative Decision II, at the bottom of Figure~\ref{fig: decisions}, shows only \textit{safe} assignments. Three \textit{easy} customers are served by the \textit{regular} technician with a working time of 210 minutes (12 $\times$ 10 + 3 $\times$ 30) and two \textit{advanced} customers are served by the \textit{expert} technician with a working time of 160 minutes (10 $\times$ 10 + 2 $\times$ 30). One \textit{easy} and one \textit{advanced} customer remain unassigned due to limited working hours and are postponed to the next period. Since this postponement exceeds the deadline of the \textit{easy} customer in the northeast, an increase in inconvenience of 1.1 is observed. However, compared to Decision I that causes no immediate increase in inconvenience, fewer (urgent) customers remain in the next state following Decision II.

\subsection{Sequential Decision Process}\label{sec: sequential}

Our problem is both stochastic and dynamic. The availability of technicians, the rework probability of \textit{risky} assignments, and the new customers in every period are stochastic. We decide every period about assignment and routing and since decisions are interconnected, the problem is dynamic. Stochastic dynamic decision problems can be modeled as sequential decision processes, defining the problem as a sequence of states, decisions, rewards (costs), revelation of information, and transitions to next states \citep{Powell2021}. As we are dealing with a minimization problem, we will use the term \enquote{costs} instead of rewards throughout the paper. In the following, we define the components of the sequential decision process. We start with preliminaries.

\subsubsection{Preliminaries.}\label{sec: preliminaries}

We assume access to a set of technicians $\mathcal{W}$, defining their skills as $b_w \in\{0,1\}$ with $b_w=0$ ($b_w=1$) representing $w$ $\in$ $\mathcal{W}$ being an \textit{regular} (\textit{expert}) technician. We define the probability of \textit{risky} assignments to remain unresolved as $p \in (0,1)$. Finally, we assume that customers can only call until period $T^c$ and a later \textit{leftover} phase is exclusively used to serve remaining incomplete customers. Notably, in contrast to other problems, the \textit{leftover} phase still follows the structure of the sequential decision process.

\subsubsection{Decision State.}\label{sec: decision state}

A decision is made in every period ($t=1,\dots,T^c, \dots,T$). Since the process only ends when all customer requests are completed, time $T$ is a random variable. A decision state \mbox{$S_t=(\mathcal{W}_t,\mathcal{K}_t,\delta_t,\rho_t,\tau_t)$} in period $t$ contains five types of information:
\begin{itemize}
    \item the set of available technicians $\mathcal{W}_t$ $\subseteq$ $\mathcal{W}$ with $m_t = |\mathcal{W}_t|$, respecting technician absences.
    \item the set of customers $\mathcal{K}_t$ with $n_t = |\mathcal{K}_t|$, containing all customer requests.
    \item the customer deadlines $\delta_t$ with $\delta_{it} \in \mathbb{N}, i=1,\dots,n_t$.
    \item the risk matrix $\rho_t$ with $\rho_{wit} \in \{0,p\}, w=1,\dots,m_t, i=1,\dots,n_t$, indicating the probability for a \textit{risky} ($\rho_{wit}=p$) and \textit{safe} assignment ($\rho_{wit}=0$) between customer $i$ and technician $w$ to remain unresolved.
    \item the travel time matrix $\tau_t$ with $\tau_{ijt} >0, i,j=0,\dots, n_t, i\neq j$, indicating the travel time between the depot $\{0\}$ and the $n_t$ customers. The travel times include the service times at customers.
\end{itemize}

\noindent We define the overall state space as $\mathcal{S}$. In the initial state defined as $S_0=(-,\emptyset,-,-,-)$, no customers have yet requested services.

\subsubsection{Decisions.}\label{sec: decisions}

In state $S_t$, a decision $x_t=(y_t,z_t) \in \mathcal{X}(S_t)$ comprises two parts. The first part, $y_t$, is the assignment of customers to technicians. The assignment variable $y_{wit} \in \{0,1\}$ is $1$ if technician $w$ visits customer $i$ in period $t$ ($0$ otherwise). The second part, $z_t$, is the routing of technicians to serve the customers. The routing variable $z_{wijt} \in \{0,1\}$ is $1$ if technician $w$ traverses arc ($i,j$) to visit $j$ coming from $i$ ($0$ otherwise). Feasible decisions satisfy the routing constraints and ensure that the working hour restrictions are respected. Consequently, our problem can be modeled as a heterogeneous team orienteering problem with time limits. For a formal definition of the decision space, we refer to Appendix~\ref{app: decision_space}.

We define $\mathcal{K}^{xu}_t = \{i \in \mathcal{K}_t \ | \ \sum\limits_{w \in \mathcal{W}_t} y_{wit} = 0 \}$ as the set of unassigned customers following decision $x_t$. The set of \textit{risky} assignments is defined as $\mathcal{K}^{xr}_t = \{i \in \mathcal{K}_t \ | \ \sum\limits_{w \in \mathcal{W}_t} y_{wit} \cdot \rho_{wit} > 0 \}$. The post-decision state $S^x_t$ when taking decision $x_t$ in state $S_t$ is defined as $S^x_t=(\mathcal{K}^{xu}_t,\mathcal{K}^{xr}_t,\delta^x_t,\rho^x_t,\tau^x_t)$ with the last three entries being the same as in the state $S_t$, but without the customers that were assigned safely by decision $x_t$.

Following the framework by \cite{Powell2021}, each decision is associated with costs. For our problem, the immediate cost $C(S_t,x_t)$ of a decision $x_t$ in state $S_t$ is the increase in inconvenience for both unassigned customers and those assigned as \textit{risky}. The value is therefore a random variable because the realized inconvenience depends on the successful completion of \textit{risky} assignments. To reflect the exponentially growing customer inconvenience caused by late services, we introduce a penalty term $\eta$ $>1$. Then, the inconvenience function $f_i(t) = \eta^{t-\delta_{it}+1}$ represents the increase in inconvenience experienced by customer $i$ during the transition from period $t$ to $t+1$, given that customer $i$ is urgent with either a due or overdue deadline ($\delta_{it}\leq t$). If the deadline is not due ($\delta_{it}>t$), the function is defined as $f_i(t)=0$. In our problem, costs are only revealed once the outcome of all \textit{risky} assignments is realized. Thus, we denote the expected costs given a state $S_t$, decision $x_t$ and probability for unresolved services $p$ as:
\begin{equation}
   \mathbb{E}\big[C(S_t,x_t)\big]= \sum\limits_{i \in \mathcal{K}^{xu}_t} f_i(t) + p \cdot \sum\limits_{i \in \mathcal{K}^{xr}_t} f_i(t). \label{eq: cost}
\end{equation}

\subsubsection{Stochastic Information and Transition Function.} \label{sec: info}

The stochastic information is threefold and defined as $\omega_{t+1}=(\mathcal{W}^{\omega}_{t+1},\mathcal{K}^\omega_{t+1}, \mathcal{K}^{r\omega}_{t+1})$. The first part is the set of available technicians, $\mathcal{W}^{\omega}_{t+1} \subseteq \mathcal{W}$. The second part $\mathcal{K}^\omega_{t+1}$ reveals a new set of customers with corresponding locations, tasks and deadlines. We recall that $\mathcal{K}^\omega_{t+1}$ is empty after the cutoff period, i.e., $t \geq T^c$. The third part relates to the realization of unresolved services, represented by the subset of \textit{risky} assignments $\mathcal{K}^{r\omega}_{t+1} \subseteq \mathcal{K}^{xr}_t$. Only now, the real cost of a decision is revealed which we define as:
\begin{equation}
c(S_t,x_t, \omega_{t+1}) = \sum\limits_{i \in \{\mathcal{K}^{xu}_t \cup \mathcal{K}^{r\omega}_{t+1}\}} f_i(t). \label{eq: real_cost}
\end{equation}

\noindent The transition function $\mathfrak{T}$ results in a new decision state in the next period $S_{t+1}=\mathfrak{T}(S_t,x_t,\omega_{t+1})=(\mathcal{W}_{t+1}^\omega,\mathcal{K}_{t+1},\delta_{t+1},\rho_{t+1},\tau_{t+1})$ with $\mathcal{K}_{t+1} = (\mathcal{K}^{xu}_t \cup \mathcal{K}^\omega_{t+1} \cup \mathcal{K}^{r\omega}_{t+1})$. The last three state entries are combined from the information of post-decision state $S^x_t$ and the set of new customers $\mathcal{K}^\omega_{t+1}$. In the special case of $\mathcal{K}_{t+1} = \emptyset \, \forall t \geq T^c$, all requests in $S_t$ were completed and the process terminates.

\subsubsection{Solution.}\label{sec: solution}

The solution of our process is a decision policy $\pi \in \Pi$, a sequence of decision rules $X^\pi_t(S_t)$ assigning a decision $x_t$ to every state $S_t$. A policy $\pi^*$ is called optimal if it minimizes the expected overall customer inconvenience for all periods starting in state $S_0$. Generally, $\pi^*$ can be defined as follows:
\begin{equation}
	\textit{$\pi^*$} = \underset{\text{{$\pi$$\in$$\Pi$}}}{\text{arg min}} \hspace{0.15cm} \mathbb{E} \Biggl[\sum_{t=1}^{T} \Big(C\big(S_t, X^\pi (S_t)\big)|S_0\Big)\Biggr].
\end{equation}

\noindent An optimal policy $\pi^*$ satisfies the Bellman equation, minimizing the sum of immediate costs plus expected future costs when following the optimal policy and being in post-decision state $S^x_t$:
\begin{equation}
	X^{\pi^*}(S_t) = \underset{x_t \in \mathcal{X}(S_t)}{\text{arg min}} \hspace{0.15cm} \mathbb{E}[C(S_t,x_t)] + \underbrace{\mathbb{E}\Bigg[\sum_{l=t+1}^{T} \Big(C\big(S_l, X^{\pi^*} (S_l)\big)|(S^x_t)\Big)\Bigg]}_{V(S^x_t)}.
\end{equation}

\noindent The second part is also known as the value function $V(S^x_t)$. It is defined as the total expected future costs when being in post-decision state $S^x_t$.

\section{Methodology}\label{sec: methodology}

In this section, we present our solution method. First, we provide a motivation and overview of our method and then introduce our state-dependent parametrization. Finally, we conclude with the algorithmic details.

\subsection{Motivation and Overview}\label{sec: motivation}

Finding effective decisions for our problem is challenging. When recalling the two decisions illustrated in the small example in Subsection~\ref{sec: example}, it is not clear which one to select. The first decision avoids immediate customer inconvenience but leaves the system in the next period congested by serving fewer customers and risking rework. Particularly in case of fewer available technicians due to absences, this leads to more urgent customers in future and thus, to higher chances for inconvenience. The second decision routes more efficiently but accepts inconvenience by not serving an urgent customer on time. Still, the resulting next state is less congested and therefore may allow for less future inconvenience. Essentially, an effective policy must carefully balance the three competing goals of (i) ensuring \textit{safe} assignments and limited risk of rework, (ii) prioritizing service for urgent customers close to or over their deadlines to avoid inconvenience, and (iii) creating efficient routes to serve many customers and reduce postponements of (urgent) customers. While considering \textit{safe} assignments is relevant in all states, the balance between routing efficiency and service to urgent customers should ideally be adapted to the current state. That is, in states with more non-urgent customers, service to isolated urgent ones may be more important, while in congested states, the focus should shift toward serving many customers efficiently. This is what we propose with our policy.

\begin{figure}[!t]
	\centering 
	\includegraphics[width = 0.75 \textwidth]{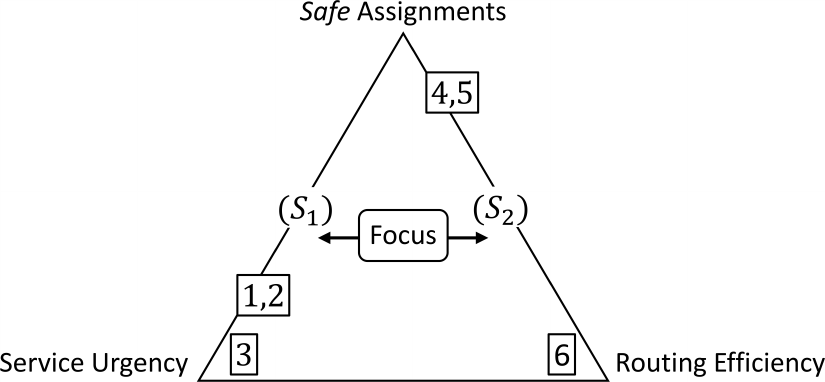}
    \caption{Illustration of the three competing goals and our policy's functionality \centering}
	\label{fig: triangle}
\end{figure}

In the following, we give an overview of the functionality of our policy. In Figure~\ref{fig: triangle}, we depict the three goals (i)-(iii) within a triangle together with an illustration of our proposed policy's functionality in the center. Our policy dynamically balances the goal focus based on state characteristics. Thus, it may position any two different states $S_1$ and $S_2$ differently within the triangle. The digits $1$-$7$ in the black boxes correspond to a numeration of benchmark policies introduced in Section~\ref{sec: benchmark}, indicating their positions within the triangle. We first discuss the three goals before demonstrating the functionality of our policy in detail.

\begin{compactitem}
    \item \textbf{\textit{Safe} assignments.} Prioritizing safe assignments means that \textit{advanced} customers are served by \textit{expert} technicians only. This will avoid any rework in future periods. However, it may prohibit consolidation opportunities for \textit{regular} technicians, e.g., in areas with several \textit{easy} and few \textit{advanced} customers. In the worst case, it leaves \textit{regular} technicians underutilized or even idling while \textit{expert} technicians travel the entire service area to serve \textit{advanced} customers. Both leads to inconvenience for existing customers and a congestion of the system.
    \item \textbf{Service urgency.} Service urgency comprises service of customers that are already delayed, but also service of customers with approaching deadlines. Focusing on deadlines only leads to similar issues as focusing on \textit{safe} assignments. Serving primarily urgent customers can help ensure that current period inconvenience is minimized but does so at the expense of untapped consolidation opportunities leaving many customers in the system and risking future congestion.
    \item \textbf{Routing efficiency.} Routing efficiency means that the limited working time of the technicians should be used to serve as many customers as possible. Therefore, assignment and routing should be conducted with respect to travel time only. Serving many customers in a period will also likely reduce the workload of future periods and avoid congestion of the system. Since customers differ in their tasks and their deadlines, this may lead to \textit{risky} assignments though. Further, few unfavorably located customers may experience significant inconvenience, especially in areas far away from the depot.
\end{compactitem}

\noindent In summary, all three individual goals have their merit. However, as we will demonstrate in Section~\ref{sec: objective}, they do not perform effectively when implemented in isolation. To avoid their shortcomings and to exploit their advantages, we propose a compromise, a method that takes all three goals into consideration in the assignment and routing decisions. In every state, it builds assignment and routing iteratively. Starting from a set of empty routes, it subsequently assigns customers to technicians and updates their routes accordingly. To select customers and technicians for assignment, we present a parametrizable score function that takes all three aforementioned goals into account. This function assesses the value of assigned customer-technician pairs in terms of the expected savings in customer inconvenience (service urgency) and additional travel time (routing efficiency). The analytical considerations of service urgency and routing efficiency within the score function are derived in the next Section~\ref{sec: score_function}. We note in advance, that the importance of \textit{safe} assignments is implicitly respected by both terms. In the framework of \cite{Powell2021}, such a method can be seen as a cost function approximation (CFA). A CFA manipulates the reward or cost function of a problem to incentivize decisions to stay flexible with regard to future developments. We adopt this idea to define a score function that manipulates the value of each assignment with respect to our goals. The basic approach is represented in Equation~\eqref{eq: score_basic}:
\begin{equation}
   \text{\enquote{Score}}= \hspace{0.1cm} (1-\alpha_{S_t}) \cdot \text{\textit{Service Urgency}} \hspace{0.1cm} - \hspace{0.1cm} \alpha_{S_t} \cdot \text{\textit{Routing Efficiency.}}
   \label{eq: score_basic}
\end{equation}

\noindent We have designed Equation~\eqref{eq: score_basic} so that high scores indicate promising assignments when creating technician tours. Specifically, assignments involving highly urgent customers (service urgency) with low additional travel time required (routing efficiency) lead to high scores. In the equation, both terms are balanced via parameter $\alpha_{S_t}$. Dependent on the magnitude of $\alpha_{S_t}$, the score function may put different emphasis on each term. To enable state-dependent balancing, we introduce a parametrization function, $\Lambda$: $\mathcal{S} \rightarrow [0,1]$, $S_t \mapsto \alpha_{S_t}$, mapping a state $S_t$ to a parameter $\alpha_{S_t}$. For any two different states $S_1,S_2$, the function $\Lambda$ determines different values for $\alpha_{S_t}$. For example, when $\alpha_{S_1} \rightarrow 0$, the score function would primarily focus on service urgency for the selection of assignments when creating tours in state $S_1$ (see Equation~\ref{eq: score_basic}). Vice versa, $\alpha_{S_2} \rightarrow 1$ would shift the focus on routing efficiency. We illustrate this distinction again in the center of Figure~\ref{fig: triangle}. Since an effective parametrization for $\alpha_{S_t}$ does depend not only on the current state and period but also on the evolution of information and decisions in future periods, we use RL to train the parametrization function $\Lambda$. Our method uses repeated offline simulations to learn the value of a parametrization function, which is then used to search for better functions.

In the next three sections, we introduce the details of our method. We first present the analytical derivation of the score function. That follows, we explain how the parametrization function $\Lambda$ is determined via RL. Eventually, we provide an overview of the full score-based assignment and routing policy, showing the integration of score and parametrization functions in an algorithmic framework.

\subsection{Analytical Derivation of Score Function} \label{sec: score_function}

Our score function respects service urgency and routing efficiency, balanced according to the current state information captured by the parameter $\alpha_{S_t}$. In this section, we evaluate properties of the Bellman equation and use these properties to motivate the design of the two parts of our score function (see Equation~\ref{eq: score_basic}).

\textit{\textbf{Service Urgency.}} The first part of the score function is motivated by the service urgency of customers. We measure service urgency by the expected \emph{increase} in inconvenience we can avoid by assigning a customer to a technician. If an assigned customers has a violated deadline, we can define these expected savings as $(1-\rho_{wit}) \cdot f_i(t)$. For \textit{safe} assignments ($\rho_{wit}=0$), savings of $f_i(t)$ are certain whereas for \textit{risky} assignments ($\rho_{wit}=p$), savings are uncertain and result in an expected value of $(1-p) \cdot f_i(t)$. For any customer with an expiring deadline in the future, this measure is indifferent as no real inconvenience can emerge. However, as we show in the following proposition, there is value in considering (artificial) savings in inconvenience also for customers with deadline periods in the future.

\begin{proposition}[Monotonicity of the value function in {$\boldsymbol{\delta_t}$}]\label{prop: monotonicity}
    Given a post-decision state $S_t^x$ with deadlines $\delta^x_t$, we construct $S_t^{x\prime}$ such that $S_t^x$ and $S_t^{x\prime}$ are identical except for their corresponding customer deadlines, i.e., $\delta^x_{it}\leq\delta_{it}^{x\prime} \hspace{0.1cm} \forall i \in (\mathcal{K}_t^{xu} \cup \mathcal{K}_t^{xr})$. Then it holds that 
    $V\big(S_t^{x^\prime}\big)\leq V\big(S_t^x\big)$.
\end{proposition}

\proof{Proof of Proposition~\ref{prop: monotonicity}.} See Appendix~\ref{app: proof}.

\begin{corollary}\label{coro: urgent_customers_first}
    Given a state $S_t$, that includes two customers $i,j \in \mathcal{K}_t$ with the same location, the same requirements, but different deadlines $\delta_{it}<\delta_{jt}$, we define two decisions $x_t^{(i)}$ and $x_t^{(j)}$. Decisions $x_t^{(i)}$ and $x_t^{(j)}$ are constructed to have the same assignment of customers to technicians and the same route except that decision $x_t^{(i)}$ includes customer $i$ and not $j$ and vice versa for decision $x_t^{(j)}$ (in both cases, the customer is assigned the same technician and their position in the route is equal). Then, it holds that $V\big(S_t^{x^{(j)}}\big)\leq V\big(S_t^{x^{(i)}}\big)$.
\end{corollary}

\proof{Proof of Corollary~\ref{coro: urgent_customers_first}.}
    Follows from Proposition~\ref{prop: monotonicity}.
\endproof

\noindent Following the corollary, even in case the deadlines of both customers $i$ and $j$ are still in the future, it is beneficial to assign customer $i$ instead of customer $j$ when $\delta_i < \delta_j$ holds. Thus, we adapt the service urgency part of our score function to consider urgency also before the deadline. The service urgency $U_{wit}$ represents the increase in inconvenience we can save for each assignment given a customer $i$ and technician $w$ in period $t$. It is a random variable as the realized (saved) inconvenience depends on the completion of \textit{risky} assignments. Thus, we can define $U_{wit}$ in expectation as:
\begin{equation}\label{eq: urgency}
\mathbb{E}\big[U_{wit}\big] = (1-\rho_{wit}) \cdot \eta^{t-\delta_{it}+1}.
\end{equation}

\noindent In Appendix~\ref{app: service_urgency}, we provide extended explanations and illustrations related to the definitions of service urgency $U_{wit}$ and inconvenience function $f_i(t)$. In particular, we show how customers with non-urgent deadlines are handled differently.

\textit{\textbf{Routing Efficiency.}} Routing efficiency is reflected by the additional travel time $\Delta\tau_t(\hat{x}_t,w,i)$, required to serve customer $i$ by technician $w$ given a current decision $\hat{x_t}$ in period $t$. However, the efficiency should not only reflect the immediate travel time required today but also the potential travel time in future periods in case of \textit{risky} assignments. For our calculation, we follow the idealized assumption that, for all future periods $t+k,k=0,1,2,\dots$, the expected required additional travel time $\mathbb{E}\big[\Delta\tau_{t+k}(\hat{x}_{t+k},w,i)\big]=\Delta\tau_t(\hat{x}_t,w,i)$ and the assignment type (\textit{safe}/\textit{risky}), i.e., $\rho_{wit}=\rho_{wit+k}$, remain the same. This means that in case of \textit{risky} assignments, the likelihood that the customer is still in the system $k$ periods in the future is $p_{wit}^k$. Under these assumptions, the immediate and future additional travel times required for each assignment, influenced by the outcomes of \textit{risky} assignments, sum up in expectation as follows:
\begin{equation}\label{eq: routing}
\mathbb{E}\big[R_{wit}\big] = \Delta\tau_t(\hat{x}_t,w,i) \cdot \sum_{k=0}^{\infty} \rho_{wit}^k = \frac{\Delta\tau_t(\hat{x}_t,w,i)}{1-\rho_{wit}}.
\end{equation}

\noindent In case of a \textit{safe} assignment ($\rho_{wit}=0$), only the additional travel time $\Delta\tau_t(\hat{x}_t,w,i)$ in period $t$ is measured which is reasonable as \textit{safely} assigned customers do not experience revisits. In case of \textit{risky} assignments, recalling that $0 < \rho_{wit}< 1$, the geometric series leads to an increase of $\Delta\tau_t(\hat{x}_t,w,i)$ by factor $\frac{1}{1-\rho_{wit}}$.

We have derived measures for service urgency and routing efficiency, each being able to capture \textit{safe} and \textit{risky} assignments. Our final score function $\zeta: (\mathcal{X}(S_t) \times [0,1] \times \mathcal{W}_t \times \mathcal{K}_t) \rightarrow \mathbb{R}, (\hat{x}_t, \alpha_{S_t}, w, i) \mapsto s$, maps, based on a given routing decision $\hat{x}_t$ and $\alpha_{S_t}$, an assignment between customer $i$ and technician $w$ on a real number (score) $s$. Inserting Equations~\eqref{eq: urgency} and \eqref{eq: routing} into \eqref{eq: score_basic} yields: 
\begin{equation}\label{eq: score_final}
s=\zeta(\hat{x}_t, \alpha_{S_t}, w, i) = (1-\alpha_{S_t}) \cdot (1-\rho_{wit}) \cdot \eta^{t-\delta_{it}+1} - \alpha_{S_t} \cdot \frac{\Delta\tau_t(\hat{x}_t,w,i)}{1-\rho_{wit}}.
\end{equation}

\subsection{State-Dependent Parametrization}\label{sec: rl}

In this section, we explain how we learn a state-dependent parametrization of our score parameter $\alpha_{S_t}$. To this end, we employ the concept of proximal policy optimization (PPO, \citealt{Schulman2017}), the current state-of-the-art extension of trust-region policy gradient methods. The state-dependent parametrization is given by a function $\Lambda$ that maps a state $S_t$ to a state-dependent parameter $\alpha_{S_t}$. We recall that $\alpha_{S_t}$ balances service urgency and routing efficiency, substantially impacting the assignments of customers to technicians in the decision-making process. We represent our problem knowledge whether $\alpha_{S_t}$ is suitable given a state $S_t$ by the continuous probability density function $\lambda(\alpha \mid S_t)$. Then, parameters $\alpha_{S_t}$, that direct decisions toward fewer accumulated costs arising from a given state $S_t$ onwards, should have a high density $\lambda(\alpha \mid S_t)$. During training, $\lambda$ is a probability density function conditioned on the current state $S_t$, from which we randomly sample parameters. However, during evaluation, we define our state-dependant parametrization function as $\Lambda(S_t)=\argmax_{\alpha}\lambda(\alpha\mid S_t)$ which represents a deterministic mapping of a state $S_t$ to a best estimated parameter $\alpha_{S_t}$. We improve $\Lambda$ by shaping $\lambda$ using offline training iterations in a RL framework. In the training framework, we do not always choose the best $\alpha_{S_t}$ according to our current problem knowledge $\lambda(\cdot \mid S_t)$ but rather sample the parameter $\alpha_{S_t}\sim\lambda(\cdot \mid S_t)$ to foster exploration.
For that purpose, we set $\lambda(\cdot\mid S_t)$ to the probability density function of the normal distribution $\mathcal{N}(\mu_t, \sigma_k)$. The mean $\mu_t$ of the normal distribution is the output of a neural network $\phi_\theta$ with parameters $\theta$ that takes state information as input. Thus, $\mu_t$ is state-dependent. The variance $\sigma_k$ of the normal distribution is interpreted as a training parameter that describes how strongly we explore in training iteration $k$. For the sake of simplicity, we represent the probability density function defined by the current network parameters $\theta$ as $\lambda_\theta$. 

In training iteration $0$, the network parameters $\theta_0$ are randomly chosen, i.e., they do not reflect any problem knowledge. However, in each training iteration $k$, we increase the likelihood of advantageous $\alpha_{S_t}$ parameters and decrease the likelihood of disadvantageous $\alpha_{S_t}$ parameters in each state by adjusting the neural network parameters $\theta_k$ accordingly. To evaluate numerically how advantageous the observed $\alpha_{S_t}$ parameters in a given state are, we define the observed advantage of having chosen a specific $\alpha_{S_t}$ in an observed state $S_t$ as $A_t$. Let $V^{\lambda_{\theta_k}}(S_t)$ be the value function, denoting the expected total costs arising from state $S_t$ onwards when sampling parameters $\alpha_{S_t}$ from the probability density function $\lambda_{\theta_k}$. Let $c_t$ denote the real immediate costs, and let $\bar{c}_t$ denote the real accumulated costs observed arising from state $S_t$ and parameters $\alpha_{S_t}$ onwards. Then, we define the advantage of parameter $\alpha_{S_t}$ chosen in state $S_t$ as $A_t=V^{\lambda_{\theta_k}}(S_t)-\bar{c}_t$. During training, we increase the density $\lambda_{\theta_k}(\alpha\mid S_t)$ if the corresponding advantage $A_t$ is positive. Vice versa, we decrease the density if the advantage is negative. However, the value function $V^{\lambda_\theta}$ is unknown and we must approximate it with the help of an auxiliary neural network $\hat{V}_{\omega_k}$ with network parameters $\omega_k$. Thus, we use $\hat{A}_t=\hat{V}_{\omega_k}(S_t)-\bar{c}_t$ as our advantage function. $\hat{V}_{\omega_k}$ is trained in parallel to $\lambda_{\theta_k}$ by minimizing an error function $L^V$ of its prediction and the observed real costs arising from a state onwards. This is not a trivial task as $\hat{V}_{\omega_{k+1}}$ might be inaccurate if we updated $\lambda_{\theta_{k+1}}$ too decisively. Therefore, the new updated probability density function $\lambda_{\theta_{k+1}}$ should ideally remain in a trust-region around the old probability density function $\lambda_{\theta_k}$ for which we believe that $\hat{V}_{\omega_{k+1}}$ is still accurate. This is the underlying concept of PPO. It ensures that updates of $\lambda_\theta$ stay in the trust region by updating the current $\theta_k$ according to
\begin{equation}
    \theta_{k+1}=\argmax_\theta \mathbb{E}_{S,\alpha\sim\lambda_{\theta_k}}\big[L(S, \alpha, \theta_k, \theta)\big], \label{eq: theta}
\end{equation}

\noindent where the objective function $L$ is defined as:
\begin{equation}
    L(S, \alpha, \theta_k, \theta)=\min\Bigg(\frac{\lambda_\theta(\alpha\mid S)}{\lambda_{\theta_k}(\alpha\mid S)}\cdot\hat{A},\text{clip}\Big(\frac{\lambda_\theta(\alpha\mid S)}{\lambda_{\theta_k}(\alpha\mid S)}, 1-\epsilon, 1+\epsilon\Big)\cdot\hat{A}\Bigg). \label{eq: rl_obj}
\end{equation}

\noindent Here, $\epsilon$ is a small parameter that determines how far the updated probability density function is allowed to deviate from the current one. We summarize the general idea of the training framework in Algorithm~\ref{alg: rl}. 

In our implementation, we represent $S_t$ as a vector of distinct features to account for the high-dimensional state variable and tackle the challenges that come along with the curses of dimensionality. Following the idea of \cite{Akkerman2022}, we have defined a set of features categorized in three groups: general state information, information on the geographical spread of customers, and information on customer deadlines. Appendix~\ref{app: feature} provides further details into the feature selection process.

\IncMargin{0.0em}
\begin{algorithm}[!t]
	\SetKwInOut{Input}{Input\,\,\,\,\,}\SetKwInOut{Output}{Output}
	\SetKw{Return}{return}
	\SetKw{break}{break}
	\Input{Value Network $\hat{V}_{\omega_0}$, Probability Density Function $\lambda_{\theta_0}$, Clip Parameter $\epsilon$}
	\Output{Network Parameter $\theta_k$}
        \For(\hspace{7.70cm} \texttt{{// Training iteration}\label{line: iterations}}){$k=0,1,2,\dots, $\label{line: learn_iter}}{\vspace{1pt}
            $D_k=\{(S, \alpha, c)\sim\lambda_{\theta_k}\}$\label{line: collect}\hspace{6.66cm}\tcp*[f]{Collect trajectories}\;
            $\hat{A}_t \leftarrow \hat{V}_{\omega_k}(S_t)-\bar{c}_t$\hspace{7.6cm}\tcp*[f]{Compute advantages}\;
            $\theta_{k+1}=\argmax_{\theta}\frac{1}{|D_k|}\sum_{(S,\alpha)\in D_k}L(S, \alpha, \theta_k, \theta)$\hspace{2.9cm}\tcp*[f]{Update policy network}\;
            $\omega_{k+1}=\argmin_{\omega}\frac{1}{|D_k|}\sum_{(S,c)\in D_k}L^V(S, \bar{c}, \omega_k, \omega)$\hspace{2.76cm}\tcp*[f]{Update value network}\;}\vspace{6pt}
        \SetNlSkip{5pt}\Return{$\theta_k$}
	\caption{Proximal Policy Optimization}
	\label{alg: rl}
\end{algorithm}

\subsection{Algorithmic Augmentation}\label{sec: impl_details}
Learning a state-dependant parametrization is immensely difficult. Even when following the original implementations of the state-of-the-art PPO algorithm as described in \cite{Schulman2017}, the final result after thousands of computationally expensive learning iterations might be unsatisfactory. As \cite{Engstrom2019} argue, algorithmic augmentations in the implementation of PPO play a major role in the success of a training run but are, if at all, only mentioned hidden in the appendix. In our preliminary tests, we came to similar conclusions. Therefore, we want to provide the community with our insights which \enquote{tricks} enabled us to learn the state-dependant parametrization when the standard implementation may not have succeeded. Inspired by \cite{Engstrom2019}, we have defined four different algorithmic augmentations.

\textit{\textbf{Scaling of Costs.}} Instead of using the actual costs for training, we scale them to the interval $[0,1]$ in order to reduce variance in training.

\textit{\textbf{Scaling of Observations.}} Instead of considering the actual observation from the environment, we scale the observation along every dimension in order to improve the quality of the learning parameter space with respect to the optimizer we apply.

\textit{\textbf{Value Function Clipping.}} We smooth the targets of the value network by considering the value loss
\begin{equation*}
L^V=\min\Big\{(\hat{V}_{\omega_k}(S_t)-\bar{c}_t)^2, \big(\text{clip}\big(\hat{V}_{\omega_k}(S_t), \hat{V}_{\omega_{k-1}}(S_t)-\epsilon, \hat{V}_{\omega_{k-1}}(S_t)+\epsilon\big)-\bar{c}_t\big)^2\Big\}
\end{equation*}
instead of the conventional mean squared error. In theory, this ensures smoother, robust updates when learning the value function and is therefore part of many PPO implementations. However, this technique's benefit is controversial, e.g., \cite{Engstrom2019} claim it has no benefit and \cite{Andrychowicz2021} argue it might even hurt performance.

\textit{\textbf{Exploration Rate.}} Instead of considering the standard deviation $\sigma_k$ as a hyperparameter that is automatically decayed, we treat it is a network parameter that is optimized over in every gradient step. This enables the network to autonomously transition between exploration and exploitation periods.

\noindent A base configuration including all augmentations is shown in the first row of Table~\ref{tab: tricks}. In configurations $(2)$-$(5)$, we respectively change one augmentation compared to $(1)$ while keeping the other three unchanged which allows us to investigate individual effects. In Section~\ref{sec: meth_ana}, we visualize the training process of each configuration and explain why we selected configuration $(4)$ for our experiments.

\begin{table}[]
\centering
\begin{tabular}{ccccc}
\hline\hline
\begin{tabular}[c]{@{}c@{}}Analyzed \\Configuration\end{tabular} & \begin{tabular}[c]{@{}c@{}}Scaling \\ of Costs\end{tabular} & \begin{tabular}[c]{@{}c@{}}Scaling\\ of Observations\end{tabular} & \begin{tabular}[c]{@{}c@{}}Value Function\\ Clipping\end{tabular}  & \begin{tabular}[c]{@{}c@{}}Exploration\\ Rate\end{tabular}\\ \hline
\\
\textit{(1)}  & \checkmark  & \checkmark  & \checkmark  & Hyperparameter    \vspace{2pt} \\
\textit{(2)}  & $\times$    & \checkmark  & \checkmark  & Hyperparameter    \vspace{2pt} \\
\textit{(3)}  & \checkmark  & $\times$    & \checkmark  & Hyperparameter    \vspace{2pt} \\
\textit{(4)}  & \checkmark  & \checkmark  & $\times$    & Hyperparameter    \vspace{2pt} \\
\textit{(5)}  & \checkmark  & \checkmark  & \checkmark  & Network Parameter \vspace{2pt} \\ \hline\hline
\end{tabular}
\vspace{0.2cm}
\caption{Different configurations based on algorithmic augmentation}
\label{tab: tricks}
\end{table}

\subsection{Score-Based Assignment and Routing Policy}\label{sec: decision_making}

As noted in Section~\ref{sec: decisions}, the decision space is a variant of a heterogeneous team orienteering problem with time limits, an NP-hard problem. Each state contains numerous customer requests and multiple technicians, making exact search mechanisms of the decision space computationally challenging. Since decisions need to be made in each state, we use a routing heuristic, $\psi: (\mathcal{X}(S_t) \times \mathcal{W}_t \times \mathcal{K}_t) \rightarrow \mathcal{X}(S_t), (\hat{x},w,i) \mapsto x$, that inserts a customer $i$ into the position within technicians $w$'s route $\hat{x}$ that causes the smallest increase in overall route duration. Thus, we expedite the solution process and obtain effective tours within reasonable runtime. For a fair comparison, we use $\psi$ for our method and all benchmark methods. Even though it is relatively straightforward, we show in Appendix~\ref{app: routing_instances} that the resulting tours are effective, and even more important, can capture different foci on service urgency, routing efficiency, and \textit{safe} assignments.

In Algorithm~\ref{alg: score_policy}, we show the conceptual procedure how decisions are derived in a state $S_t$. The state variable $S_t$ and a parametrization function $\Lambda$ serve as the input parameters. The final decision $x^*_t$ represents the output, determining the technician routing for that state. A (feasible) starting solution is a decision $x_t$ that does not assign any customers, i.e., empty routes for all technicians from the depot to depot ([0,0]). Then, customers are added subsequently to the routing solution $x_t$ as follows. From Line~\ref{line: candidates} to \ref{line: candidate_end}, we iteratively assign every remaining customer $i$ to every technician $w$ as we determine the score $s$ of that assignment via our score function $\zeta$, and update the (preliminary) routing decision $\overline{x_t}$ with function $\psi$. If the score value is higher than all previously observed values and the decision is feasible (Line~\ref{line: feas}), the decision is stored (Line~\ref{line: best_score}-\ref{line: best_route}). Once all score values are calculated, the routing decision $x^*_t$ is updated according to the assignment with the highest score value (Line~\ref{line: update}). The process is repeated until either all customers are assigned or no feasible assignments remain.
\IncMargin{2.2em}
\begin{algorithm}[!t]
	\SetKwInOut{Input}{Input\,\,\,\,\,}\SetKwInOut{Output}{Output}
	\SetKw{Return}{return}
	\SetKw{break}{break}
	\Input{State Variable $S_t$, Parametrization Function $\Lambda$}
	\Output{Technician Routing $x^*_t$}
    $x^*_t$, $x_t$ $\leftarrow$ ([0,0],\dots,[0,0])\hspace{4.36cm}\label{line: init_tours}\tcp*[f] {Set default routing decision}\;
    $\alpha_{S_t}$ $\leftarrow$ $\Lambda(S_t)$ \hspace{5.59cm}\label{line: param}\tcp*[f] {Determine $\alpha_{S_t}$ for current state}\;
    \While(\hspace{4.50cm}\texttt{{// Start iterative assignment process}\label{line: main_loop}}){$\mathcal{K}_t$ $\neq \emptyset$\label{line: while_loop}}{\vspace{1pt}
        $\hat{x}_t$ $\leftarrow$ $x_t$\hspace{8.24cm}\label{line: init_x}\tcp*[f]{Copy current routes}\;
        $s^*$ $\leftarrow$ $\infty$\hspace{7.40cm}\label{line: init_c}\tcp*[f]{Set default score value}\;
        \ForAll(\hspace{3.53cm} \texttt{{// Evaluate assignment candidate}\label{line: candidates}}){$i$ $\in$ $\mathcal{K}_t$, $w$ $\in$ $\mathcal{W}$}{\vspace{1pt}
            $s$ $\leftarrow$ $\zeta$($\hat{x}_t, \alpha_{S_t}, w, i$\big)\label{line: score}\hspace{4.50cm}\tcp*[f]{Calculate assignment score}\;
            $\overline{x_t}$ $\leftarrow$ $\psi$\big($\hat{x}_t, w, i$\big)\label{line: pre_insert} \hspace{2.99cm}\tcp*[f]{Create preliminary routing decision}\;
            \If(\hspace{1.87cm} \texttt{{// Check best score and feasibility}\label{line: feas}}){$s$ $<$ $s^*$ $\land$ $\overline{x}_t$ $\in$ $\mathcal{X}(S_t)$}{\vspace{1pt}
                $s^*$ $\leftarrow$ $s$\label{line: best_score}\hspace{6.17cm}\tcp*[f]{Store lowest score value}\;
                $i^*$ $\leftarrow$ $i$\label{line: best_cust}\hspace{6.09cm}\tcp*[f]{Store "cheapest" customer}\;
                $x_t$ $\leftarrow$ $\overline{x}_t$\label{line: best_route}\hspace{4.60cm}\tcp*[f]{Store cheapest routing decision}\;
                }}\vspace{1pt}\label{line: candidate_end}
        \If(\hspace{4.03cm}\texttt{{// Check if feasible assignment exists}\label{line: feas_int}}){$s^*$ $\neq$ $\infty$}{\vspace{1pt}
            $\mathcal{K}_t$ $\leftarrow$ $\mathcal{K}_t$\textbackslash\{$i^*$\}\label{line: remove}\hspace{7.41cm}\tcp*[f]{Remove customer}\;
            $x^*_t$ $\leftarrow$ $x_t$ \label{line: update}\hspace{6.77cm}\tcp*[f]{Update routing decision}\;\vspace{0pt}}\label{line: best_end}
        \Else(\hspace{6.88cm}\texttt{{// No feasible assignment exists}\label{line: terminate}}){\vspace{1pt}
            \break\label{line: finish}\vspace{0pt}\tcp*[f]{Terminate algorithm}}\vspace{0pt}}
    \SetNlSkip{4.5pt}\Return{$x^*_t$}
	\caption{Score-Based Routing and Assignment Policy}
	\label{alg: score_policy}
\end{algorithm}

\section{Computational Evaluation}\label{sec: evaluation}

In this section, we first describe the test instances and benchmark policies. Then, we present the computational analysis, providing insights into the objective value and examining both the methodology and the problem.

\subsection{Test Instances}\label{sec: instances}

In our experiments, the fleet consists of three \textit{regular} and three \textit{expert} technicians. We assume an absence rate of 10\% per technician and day \citep{IWD2024}. Technicians do not work more than seven hours a day in the field. We assume on-site service times of 30 minutes. Following the setting of the companies discussed in the introduction, customers are uniformly distributed across a quadratic service area of 200 $\times$ 200 kilometers, with the depot located in the center. We further assume Euclidean distances and an average driving speed of 60 kilometers per hour to capture the road network and a mix of highways, rural roads, and cities. We assume operations of a month with orders coming in the first three weeks (five working days per week) and a subsequent \textit{leftover} phase. The cutoff period is therefore $T^c=16$.

We expect 180 customer requests during a week leading to an expected number of six customer requests per technician and workday. Yet, as service requests accumulate over the weekend, we assume that the expected number of requests on Mondays is three times as high as on the other days. Technically, the expected number of (new) daily customer requests, excluding Mondays, follows the normal distribution $\mathcal{N}(\mu, \sigma^2)$ with $\mu = \frac{180}{7} \approx 25.7$, coefficient of variation $cv = \frac{1}{6}$ and $\sigma = cv \times \mu \approx 4.3$. On Monday, the number is tripled.

We assume a time span of two days after customer requests are revealed during which no inconvenience arises. Thus, a service provider has three periods to serve a request on time. After that, we assume an increase in inconvenience of 10\% from period to period (see Figure~\ref{fig: functions}). The likelihood of \textit{easy} and \textit{advanced} tasks is even, matching the skill distribution within the technician fleet. \textit{Risky} assignments remain unresolved in half the cases, i.e., with probability $p=0.5$. Based on these parameters, we create 150 test instances, i.e., realizations of the entire sequential decision process, for evaluating our method.

\subsection{Benchmark Policies}\label{sec: benchmark}

We compare our method, which we call \textit{DB} (dynamic-balance) policy, to six problem-oriented benchmark policies and one method-oriented policy. All policies follow the general procedure proposed in Algorithm~\ref{alg: score_policy}, relying on routing heuristic $\psi$, but with different and static foci on goal dimensions in the score function. According to the following numbering $1$-$7$, we have positioned each policy in Figure~\ref{fig: triangle}. The first three policies follow different ideas of a myopic approach to minimize the immediate increase in customer inconvenience (service urgency). The next three policies entirely disregard the increase in customer inconvenience and direct their focus on several assignment and routing rules. The last policy is method-oriented to highlight the value of both balancing goal dimensions and using RL for state-dependent tuning of the score function as performed in our \textit{DB} policy.
\begin{compactenum}
    \item \textbf{\textit{MYSF}}: This myopic-safe policy assigns customers with the longest expired deadlines first as they contribute most to the increase in inconvenience when not visited. Thereby, it allows \textit{expert} technicians to visit all customers but no \textit{regular} technicians to perform \textit{advanced} tasks. 
    \item \textbf{\textit{MYEX}}: This myopic-exclusive policy assigns customers with the longest expired deadlines first. Thereby, the policy splits the workforce. \textit{Expert} technicians only perform \textit{advanced} tasks and \textit{regular} technicians only perform \textit{easy} tasks. 
    \item \textbf{\textit{MYEF}}: This myopic-efficient policy assigns customers with the longest expired deadlines first. Thereby, the policy assigns based on travel time increase only and disregards the risk of mismatches.
    \item \textbf{\textit{SF}}: Identical to \textit{MYSF}, but it disregards the inconvenience of customers when assigning.
    \item \textbf{\textit{EX}}: Identical to \textit{MYEX}, but it disregards the inconvenience of customers when assigning.
    \item \textbf{\textit{EF}}: Identical to \textit{MYEF}, but it disregards the inconvenience of customers when assigning.
    \item \textbf{\textit{SB}}: This static-balance policy is designed to show the impact of balancing goal dimensions and our state-dependent parametrization. To this end, this policy applies our score function with static parameter $\alpha$. The optimal parameter is determined via enumeration. Details are presented in Appendix~\ref{app: methodological}. 
\end{compactenum}

\subsection{Objective Value and Average Delay}\label{sec: objective}

\begin{figure}[!t]
    \centering 
    \includegraphics[width = 0.56 \textwidth]{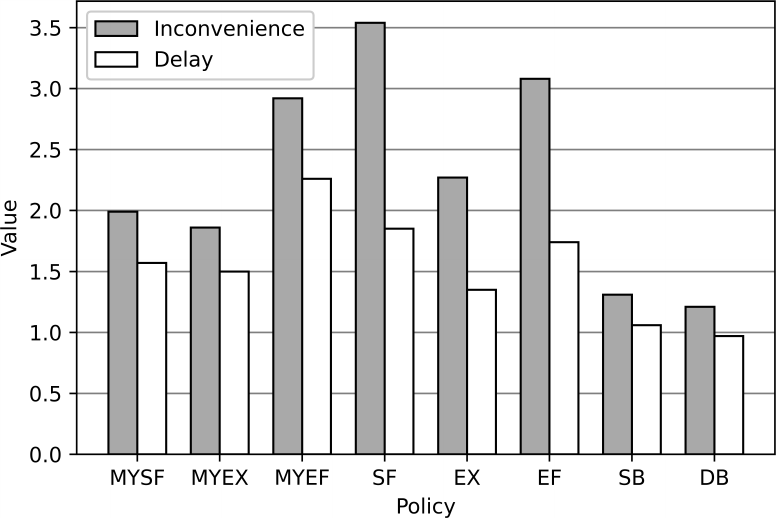}
    \caption{Average inconvenience and average delay per customer \centering}
    \label{fig: objective}
\end{figure}

First, we compare the objective values of all policies. The grey bars of Figure~\ref{fig: objective} show the average inconvenience per customer. Each bar on the x-axis represents a policy, the y-axis shows the respective inconvenience value. For a detailed analysis related to the development of inconvenience over time, we refer to Appendix~\ref{app: cummulative_inconvenience}.

We observe significant differences in inconvenience. Policies \textit{SB} and \textit{DB} perform substantially better as they induce the fewest average customer inconvenience. The myopic policies \textit{MYSF, MYEX, MYEF} perform better than their corresponding policies \textit{SF, EX, EF} that do not consider customer deadlines. As expected, there is value in considering customer deadlines when aiming to minimize customer inconvenience. From the policies considering deadlines, policy \textit{MYEF} performs worst. This policy ignores heterogeneity in tasks and workforce, and only aims for the most efficient routing. This leads to many \textit{risky} assignments and rework and eventually to more inconvenience for future customers. 

Interestingly, the \textit{MYSF} policy performs worse than \textit{MYEX}. \textit{MYSF} avoids any \textit{risky} assignments but allows assigning \textit{easy} tasks to any technician. This leads to a backlog of the \textit{expert} technicians and many unserved \textit{advanced} tasks, while \textit{regular} technicians remain idle. Consequently, the \textit{MYEX} policy performs better. This policy splits the workforce, only allowing \textit{easy-regular} and \textit{advanced-expert} assignments. This distributes the workload more equally among the technician and avoids congestion of any type of task. The  difference between  \textit{MYSF} and \textit{MYEX} indicates that a company might rather split the customers and workforce than combining them in a myopic way. Finally, we observe that our balanced approaches, \textit{SB} and \textit{DB}, outperform all other policies by a significant amount. While a static balance \textit{SB} is already beneficial, a dynamic balance \textit{DB} further increases performance by almost 8$\%$. Besides the average inconvenience, Figure~\ref{fig: objective} also shows the average delay per customer in days via the white bars. The delay values of the different policies are smaller, as expected due to the definition of the inconvenience function. However, for the \textit{SF}, \textit{EX} and \textit{EF} policies, the reductions are significantly greater. These policies serve more customers on-time, but leave a few customers in remote area experiencing high delays (see Figure~\ref{fig: geo_delay}). Due to the non-linearity of the inconvenience function, these delayed customers contribute disproportionately to the overall inconvenience.

\subsection{Analyzing the State-Dependent Parametrization}\label{sec: meth_ana}

In this section, we first discuss the performance of augmentation details that we have implemented during the learning process. Then, we analyze the impact of selective features on the parametrization function $\Lambda$. 

\begin{figure}[!t]
    \centering 
    \includegraphics[width = 0.56 \textwidth]{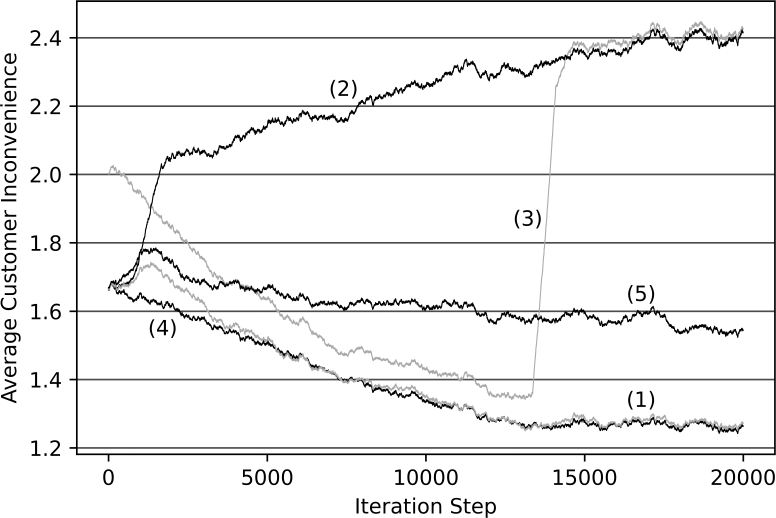}
    \caption{Learning curves for five algorithmic configurations (see Table~\ref{tab: results_app}) \centering}
    \label{fig: rl_training}
\end{figure}

\cite{Engstrom2019} list different algorithmic augmentations that are usually only explained in the appendix but are attributed to successful PPO implementations. Figure~\ref{fig: rl_training} shows the learning processes for five configurations (1)-(5) introduced in Table~\ref{tab: tricks}, each including different augmentations in the implementation. The x-axis shows the iteration step, and the y-axis the average customer inconvenience. We provide explanations for each graph in the subsequent numeration. Thereby, the numbers corresponds to the respective configuration visualized in Figure~\ref{fig: rl_training}. We recall that configuration (4) reflects our \textit{DB} policy, which we have used for the computational evaluation.

\begin{compactenum}
\item[(1)] Incorporating value function clipping leads to slightly slower learning in the beginning but ultimately converges to a policy that is marginally worse than our selected policy. We do not observe any benefit to value function clipping.
\item[(2)] Without scaling of costs, the algorithm diverges due to the high variance in observed costs.
\item[(3)] Leaving out normalization of observations hampers the learning process in the beginning. An edge case in a later iteration causes gradients to explode, introducing a kink in the learning curve. The policy diverges from this point on. We observed similar behaviors for differently seeded training runs.
\item[(4)] Our chosen configuration (scaling of costs, normalization of observations, no value function clipping, decaying the standard deviation) is characterized by a smooth learning curve that converges after 15,000 iterations at an average inconvenience level that is smaller than the result of the other configurations.
\item[(5)] Enabling our policy to autonomously learn the standard deviation $\sigma_k$, responsible for balancing exploration and exploitation, leads to a very slow training progress. In an extended simulation, we observed convergence after around 100,000 iteration steps (see Figure~\ref{fig: rl_curve_complete} in Appendix~\ref{app: long_training}). Eventually, this policy shows even slightly better performance during evaluation, about a 2\% improvement compared to our \textit{DB} policy and a 10\% improvement compared to the \textit{SB} policy. It might therefore be a valid strategy for problems where training times are not an issue, e.g., with small decision spaces. 
\end{compactenum}

\noindent To explain the output of our trained function $\Lambda$, we use \enquote{Shapley Additive Explanations} (SHAP), an approach derived from game theory. We have extracted selected features that either demonstrate a significant influence on $\alpha_{S_t}$ or are pertinent to our problem analysis (see Table~\ref{tab: feauture_selection}). We emphasize though that the final parametrization for $\alpha_{S_t}$ is the result of the interplay of all features and considering them in isolation should be done with care. The idea of our analysis is to show the impact on $\alpha_{S_t}$ if a feature value is smaller/larger than expected. To this end, we first take the mean of all feature values for all test instances. Then, for feature values below and above this mean, we calculate the respective mean output values for $\alpha_{S_t}$. Finally, we calculate the percentage change related to the overall mean $\alpha_{S_t}$ for the test instances.

In the first column of Table~\ref{tab: feauture_selection}, we present our selected features. The second (third) column shows the impact on $\alpha_{S_t}$ for states where the feature value is below (above) the mean. Due to the structure of our score function (see Equation~\ref{eq: score_final}), smaller values for $\alpha_{S_t}$ emphasize the immediate savings in customer inconvenience (service urgency) while higher values indicate a stronger emphasis on routing efficiency. The first two features consider the distances between customer locations and the depot for \textit{easy} and \textit{advanced} customers, respectively. For \textit{easy} customers, we see a significant increase in $\alpha_{S_t}$ values of almost 20$\%$ when their locations are farther away from the depot. As visiting these customers is associated with higher expected travel times, the policy prioritizes efficient routing for two reasons:
First, efficient routing is required to serve many far away customers. Second, in case technicians are already in that area, they should serve both urgent and less urgent customers. Conversely, for \textit{easy} customers located close to the depot, the policy decreases $\alpha_{S_t}$ values by 11$\%$. Travel times typically decrease, enabling more customer visits even with reduced emphasis on routing efficiency. This shift in focus allows for a closer consideration of minimizing customer inconvenience. Compared to \textit{easy} customers, the distance from the depot to \textit{advanced} customers also influences the value of $\alpha_{S_t}$ in a similar way, though the effect is significantly smaller. These differences may be explained by the risk of rework \textit{advanced} customers represent when visited by a \textit{regular} technician. This risk is captured not only in the efficiency but also in the inconvenience part of the score function potentially explaining the more balanced selection of $\alpha_{S_t}$.

\begin{table}[]
\centering
\begin{tabular}{lcc}
\hline
\vspace{1pt}
\multirow{2}{*}{\begin{tabular}[c]{@{}l@{}}  \\ State Feature\end{tabular}} & \multicolumn{2}{c}{\begin{tabular}[c]{@{}c@{}} \\ Impact on $\alpha_{S_t}$ (\%)\end{tabular}} \\ \cline{2-3} 
\multicolumn{1}{c}{} & \begin{tabular}[c]{@{}c@{}} \\ Low Feature Value\end{tabular} & \begin{tabular}[c]{@{}c@{}}  \\ High Feature Value\end{tabular} \\ \hline  \\
Distance for \textit{easy} customers to depot      & \hspace{0cm}   -11.0 & \hspace{0cm}    +19.1 \vspace{2pt} \\ 
Distance for \textit{advanced} customers to depot  & \hspace{0.2cm} -2.2  & \hspace{0.2cm}  +3.1  \vspace{2pt} \\
\textit{Easy} customers with overdue deadline      & \hspace{0.2cm} -8.2  & \hspace{0cm}    +11.6 \vspace{2pt} \\
\textit{Advanced} customers with overdue deadline  & \hspace{0cm}   -11.2 & \hspace{0cm}    +12.0 \vspace{2pt} \\
Number of available technicians                    & \hspace{0.2cm} -0.8  & \hspace{0.2cm}  +0.7  \vspace{2pt} \\ \hline\hline
\end{tabular}
\vspace{0.2cm}
\caption{Influence of (selected) features on parameter $\mathbf{\boldsymbol{\alpha}_{S_t}}$}
\label{tab: feauture_selection}
\end{table}

Next, we analyze features considering the number of \textit{easy} and \textit{advanced} customers with overdue deadlines (see Table~\ref{tab: feauture_selection}). These customers are characterized by the fact that their deadlines are due in the current decision state or are already overdue ($t \geq \delta_{t}$). Here, we observe similar patterns for both types of customers. We can see that, for many customers with overdue deadlines, our policy increases $\alpha_{S_t}$ to improve the routing efficiency (+11.6$\%$ and +12.0$\%$). As all these customers would contribute to the increase in customer inconvenience, our policy evaluates it beneficial to increase the number of visits and reduce system congestion. The opposite is true for fewer customers. Then, the policy particularly emphasizes the inconvenience to complete service for the few customers with overdue deadlines.

The last row presented in Table~\ref{tab: feauture_selection} considers the absences of technicians. We see no significant change of the $\alpha_{S_t}$ parameter related to the number of available technicians. This indicates that the general functionality of our policy remains the same independent of the number of technicians in the system. Even when technicians are absent, an additional focus either on routing efficiency or inconvenience does not add much benefit.

\subsection{Problem Analysis}\label{sec: problem_ana}

In this section, we analyze the impact of our policy, in particular, the difference to conventional (myopic) decision making. Thus, we mainly focus on the three benchmark policies \textit{MYSF}, \textit{MYEX} and \textit{MYEF}. In Appendix~\ref{app: kpi}, we present extended metrics and compare the performance of all benchmark policies.

\begin{figure}[!t]
    \centering
    \begin{minipage}{0.49\textwidth}
    \centering
    \includegraphics[width=\textwidth]{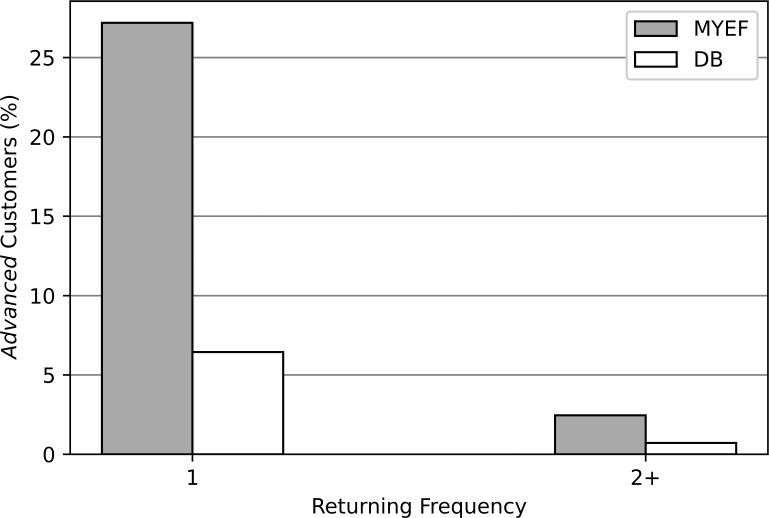}
    \caption{Frequencies of unresolved services \centering}
    \label{fig: rework_distribution}        
    \end{minipage}\hspace{0.2cm}
    \begin{minipage}{0.49\textwidth}
    \centering
    \includegraphics[width=\textwidth]{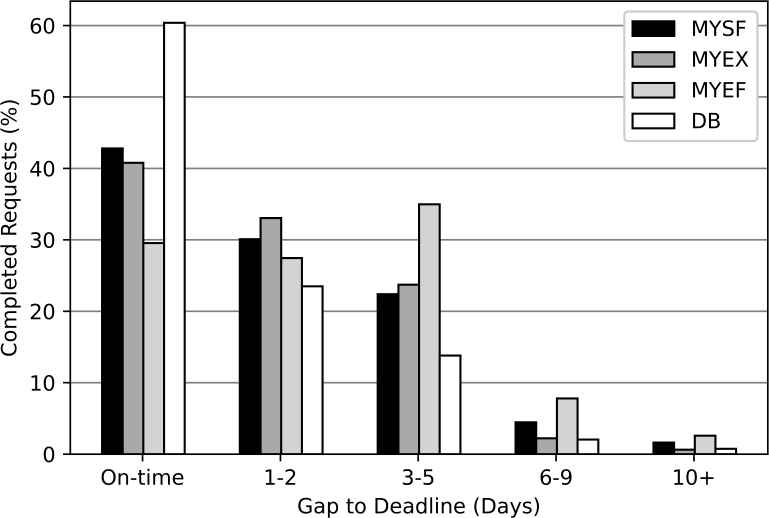}
    \caption{Temporal service completions \centering}
    \label{fig: service_com}
    \end{minipage}
\end{figure}

While we selected inconvenience as the main objective of this work, there are other important key performance indicates. In our experiments, we assume that the main inconvenience stems from the waiting time until a task is completed. However, there is also an inconvenience associated with repeated technician visits. Thus, we investigate the percentages of times customers are visited more than once. In Figure~\ref{fig: rework_distribution}, we visualize the percentage of returning visits observed for \textit{advanced} customers due to \textit{risky} assignments. The x-axis show the number of revisits. On the y-axis, we show the share of \textit{advanced} customers having experienced repeated visits. By definition, the \textit{MYSF} and \textit{MYEX} policies exclude \textit{risky} assignments and thus, do not induce rework. For any policy that disregards skills, we expect that 25\% of all \textit{advanced} customers experience a repeated visit, given that their likelihood of being assigned as \textit{risky} is 50\%, and the likelihood of unresolved services is also 50\%. However, with a \textit{MYEF} policy, around 27\% of all \textit{advanced} customers are revisited once and another 2.5\% twice or more. Thus, over time, \textit{advanced} customers outnumber \textit{easy} ones as they are postponed more often due to unresolved services, leading to more \textit{risky} assignments and returning visits. Conversely, a \textit{DB} policy results in a remarkably small number of revisits of around $7\%$, with only a few customers being visited more than twice. This indicates that the \textit{DB} uses the option of \textit{risky} assignments carefully, but effectively, as previous results have shown.

In Figure~\ref{fig: service_com}, we analyze the service completions among all customers related to their deadlines. The x-axis represents the differences in days between service completion and individual deadline period. The y-axis represents the share of all completed requests. The \textit{DB} policy serves about 60\% of all customers on time, whereas the myopic policies show rates between 30\% to 45\%. Further, the \textit{DB} policy keeps especially the number of customers with higher delays of more than three days on a small level. Thus, our policy does not improve average objective values at the expense of a few customers. This rather \enquote{fair} behavior can also be observed when analyzing the regional spread of inconvenience in the next section. 

\begin{figure}[!t]
    \centering 
    \includegraphics[width = 1.0 \textwidth]{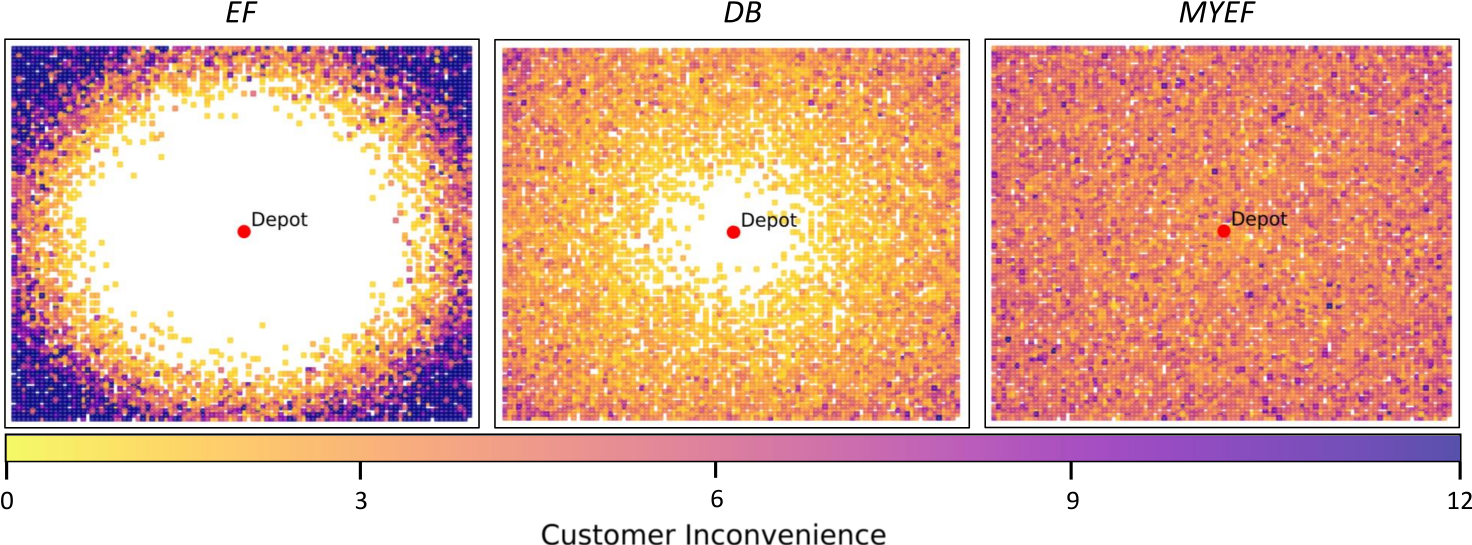}
    \caption{Regional spread of average customer inconvenience \centering}
    \label{fig: geo_delay}
\end{figure}

Figure~\ref{fig: geo_delay} shows the service area with the depot located in the center for the \textit{EF}, \textit{DB} and \textit{MYEF} policies. We include benchmark policy (\textit{EF}) to visualize the impact of focusing solely on routing efficiency. Each dot represents a specific (x,y) coordinate with the color indicating the average observed inconvenience for all customers having requested from this location (white areas indicate no customer inconvenience). The left panel shows the \textit{EF} policy that serves all customers on-time who are located near the depot. However, toward the edges, customer inconvenience increases significantly. As this policy prioritizes routing efficiency, customers located in remote regions are completely overlooked across multiple periods. This characteristic resembles the ALP policy introduced in \cite{Khorasanian2024}, rejecting referrals beyond a certain distance from the depot and primarily accepting referrals near the depot. This approach leads to poor performances in our problem as we cannot reject, but only postpone requests. For the \textit{MYEF} policy shown in the right panel, we see an almost equally distributed inconvenience spread across the entire service area. As this policy prioritizes serving the most urgent customers over considering associated travel times, all customers experience similar inconveniences. While this might be fair, the average performance per customer is poor. Similar to \textit{EF}, \textit{DB} (center panel) is able to prevent any inconvenience for customers located close to the depot. Still, it manages to keep the inconvenience levels low even for customers distant from the depot. It ensures that that customers in remote regions are not overlooked, preventing them from experiencing significant service delays and inconveniences.

\begin{figure}[!t]
    \centering
    \begin{minipage}{0.478\textwidth}
        \centering
        \includegraphics[width=\textwidth]{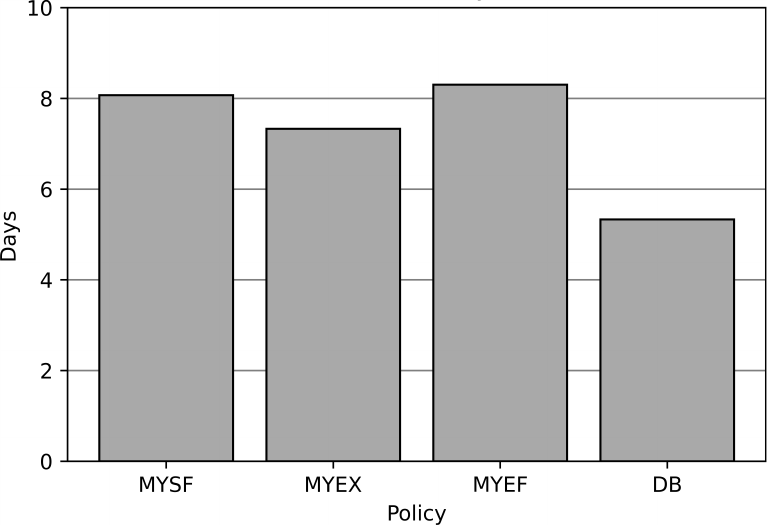}
        \caption{Duration of \textit{leftover} phase \centering}
        \label{fig: leftover}
    \end{minipage}\hspace{0.2cm}
    \begin{minipage}{0.49\textwidth}
        \centering
        \includegraphics[width=\textwidth]{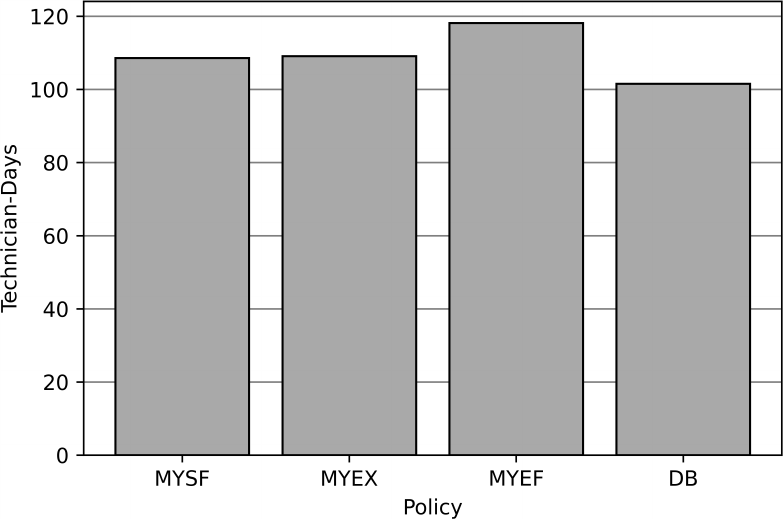}
        \caption{Required workforce resources \centering}
        \label{fig: resources}
    \end{minipage}
\end{figure}

We will now analyze how our policy improves upon the required time to serve all customers in the \textit{leftover} phase. In Figure~\ref{fig: leftover}, we depict the duration of the \textit{leftover} phase which is required to complete all remaining requests after period $T^c$. The policies are shown on the x-axis, the duration in days on the y-axis. With a \textit{leftover} phase of around five days, our \textit{DB} requires 30\% to 40\% less time to complete all remaining requests compared to myopic policies, which need around eight additional days. Thus, in contrast to the benchmark policies, \textit{DB} is able to accept new requests again after around one week, i.e., at the beginning of the next month. Moreover, \textit{DB} requires the fewest total amount of working times (travel times plus service times) to complete all requests. In Figure~\ref{fig: resources}, the x-axis and y-axis represent the policies and required technician-days, respectively. Thereby, one technician-day corresponds to seven hours, equivalent to an entire working day. Figure~\ref{fig: resources} illustrates that the \textit{DB} policy needs 7\% fewer technicians-days compared to the \textit{MYSF} and \textit{MYEX} and even 20\% fewer than \textit{MYEF}.

Ultimately, based on the results shown, our policy brings improvements across three key areas: (i) enhanced service quality for customers with reduced inconvenience and rework, (ii) increased satisfaction among technicians who successfully resolve over 90\% of their assigned tasks during the first visit, and (iii) improved operational performance for company by optimizing resource utilization.

\subsection{Workforce Analysis}

In this final section, we want to elaborate on the workforce. We investigate the implications of both different absence rates and heterogeneously qualified technician fleets on policy performance. 

\begin{figure}[!t]
    \centering
    \begin{minipage}{0.488\textwidth}
        \centering
        \includegraphics[width=1.0\textwidth]{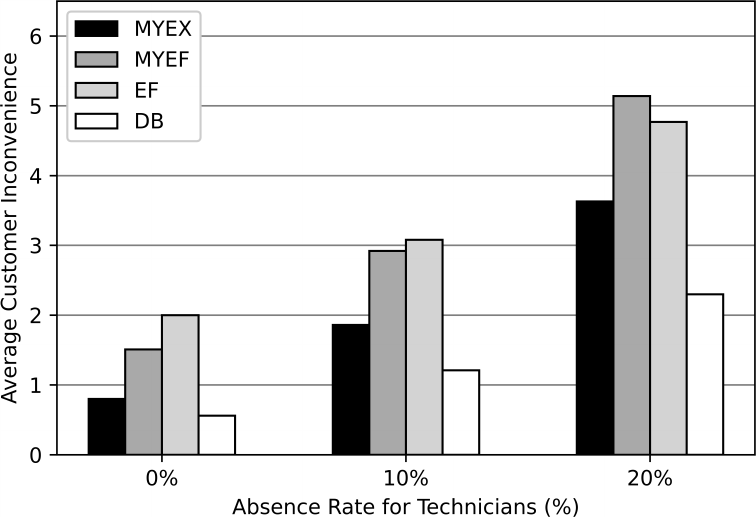}
        \caption{Impact of absence rate \centering}
        \label{fig: absences}
    \end{minipage}\hspace{0.2cm}
    \begin{minipage}{0.488\textwidth}
        \centering
        \includegraphics[width=\textwidth]{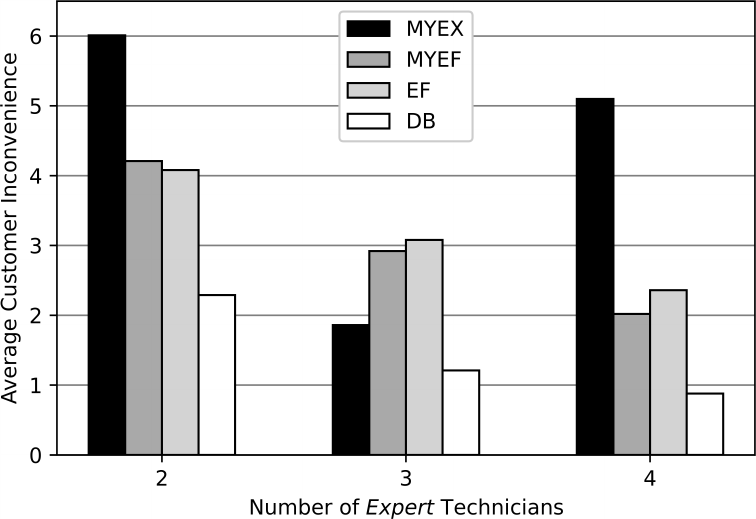}
        \caption{Impact of fleet skill \centering}
        \label{fig: experts}
    \end{minipage}
\end{figure}

On the x-axis in Figure~\ref{fig: absences}, we show various rates indicating the daily probability of each technician being absent, for instance, due to illness. The average customer inconvenience is shown on the y-axis. With every technician being certainly available each day (0\%), inconvenience rates remain on a relatively small level. As absence rates increase, we recognize a clear increase in inconvenience. An absence rate of 20\% implies more than one technician less on expectation assuming a fleet size of six technicians. The resulting higher workload leads to more customer postponements and long waiting customers with high inconvenience rates. We observe that \textit{DB} outperforms the benchmark policies regardless of the absence rate. It even achieves similar results with higher rates (20$\%$) compared to benchmark policies with lower rates (10$\%$). Comparing the policies \textit{MYEF} and \textit{EF}, we see that, with less available technicians, routing efficiency becomes more important to reduce the increase in inconvenience. Decision states become highly congested with many late customers, demonstrating that numerous visits may be an effective strategy for these occasions. We see this for absence rate of 20\%, when \textit{EF} outperforms \textit{MYEF}. In Figure~\ref{fig: experts}, we illustrate the performance across different skill distributions within the technician fleet. The x-axis shows the number of \textit{expert} technicians, the y-axis the average inconvenience. Due to its definition, \textit{MYEX} performs best when the proportion of \textit{easy} and \textit{advanced} customers is equal (50\%)  to the proportion of \textit{regular} and \textit{expert} technicians as this avoids idle technicians. All other policies perform better with a more skilled fleet which is intuitive as \textit{advanced} customers experience less rework and consequently, less inconvenience. We further observe that \textit{DB} induces less inconvenience with only three \textit{expert} technicians than benchmark policies with four \textit{expert} technicians.

\section{Conclusion}\label{sec: conclusion}

In this work, we have addressed the problem of dynamically assigning tasks to technicians with varying qualifications. We have shown that an isolated focus on individual goal dimensions leads to very ineffective solutions. Myopically minimizing the customer inconvenience (service urgency) is resource-demanding and congests the system in the long term. Maximizing the number of visited customers (routing efficiency) leads to mismatches and overlooks customers in remote areas who experience long waiting times. Focusing on \enquote{perfect} matches lead to limited visits per day and long waiting times for customers. Consequently, we have developed a method that takes a combination of the goal dimensions into account. Steered by a state-dependent parametrization, we could improve the performance across various dimensions. Our method completes requests faster and more reliably, while requiring fewer resources, which is advantageous for both customers and business operations. As we have used reinforcement learning to train our state-dependent parametrization, we have investigated the impact of different state features for decision-making. In states where customers are located farther away from the depot or many customers' deadlines are overdue, decisions need to focus on creating efficient routes instead of serving the most urgent customers and vice versa. In addition, we have pointed out that algorithmic augmentations must be carefully selected and considered when using reinforcement learning for state-dependent parametrizations in general.

There are several avenues for future work. In our work, we assumed two types of technicians and tasks. Future work may investigate both parts in more details. For example, technicians might learn based on their assigned works as discussed in \cite{Chen2016}. Consequently, operational assignments might be paired with strategical workforce development. At the same time, the set of tasks may be considered in more details. Instead of two groups of tasks, technicians may face a set of different tasks and each technician may have different skills with respect to the individual tasks. Furthermore, while we assumed that the task difficulty is known (e.g., due to diagnostics), there might be cases where the task difficulty only reveals when the customer is visited. Here, it might be valuable to have some \textit{regular} technicians \enquote{scouting} tasks with uncertain skill requirements \citep{vanMoeseke2022}. The focus of our work has been the risk of rework due to technician skills while the literature has focused on rework due to missing spare parts \citep{Pham2024}. Future research may combine both sources when considering both skills and parts in an integrated fashion. Another extension could consider optimizing the distribution of workload equally among technician. We have examined significant disparities in capacity utilization among technicians, particularly in policies that avoid assignments prone to rework, which ultimately results in poorer overall performance. Improving equitable workload allocation, such as balancing the demand served per technician, may not only enhance service performance but increase acceptance and morale among the workforce \citep{Matl2018}. Related to this point, practice has shown that technician absences are often related to a stressful work environment. Here, the interdependence between planning and potential absences may deserve further attention. Furthermore, the developed method shows how reinforcement learning can be used to tune interpretable and anticipatory policies. Future work may transfer the general concept to other problem domains where state-dependent tuning is required. 

\newpage

\begin{appendices}\label{sec: appendix}

\noindent In the Appendix, we first present an extensive literature review. We then provide a formal definition of the problem's decision space, followed by extended methodological details and additional results.
 
\section{Extensive Literature Review}\label{app: literature}

In this section, we discuss related literature on technician routing, multi-period service routing, and routing with repeated customer visits. A summarized list of references can be found at the end of the appendix.

\subsection{Deterministic Technician Routing}

\cite{Pahl2011} introduce an extended vehicle routing problem that incorporates skill levels. This work has offered promising initial insights for subsequent solution approaches that emerged in the field. The literature classifies these types of problems as technician routing and scheduling problems (TRSP). They consider different qualified technicians with limited available working hours who complete on-site tasks that have certain service requirements \citep{Pourjavad2022}.

\cite{Braekers2016} analyze the trade-off between operational costs and client satisfaction in a home and health care routing and scheduling problem. Nurses have different levels of qualifications, allowing them to perform jobs only at clients requiring their qualifications. The authors show that the average service level can be significantly improved with a relatively small additional increase in costs. \cite{Schwarze2015} move the focus of cost-oriented to time-related objectives in TRSP and add time windows as further restrictions in the context of an airport ground control. It turns out that multi-objective approaches are worthwhile to examine as possibilities of alternative objectives are better exploited. They recommend to implement heuristic solution methods to approach large instances in TRSP with time windows. \cite{Kovacs2012} introduce a TRSP in the field of infrastructure and maintenance service. As an extension, they incorporate jobs that can only be completed by a group of technicians. An implemented adaptive large neighborhood search computes high solution qualities with cost decreases of 10\% within short computational runtime. Further applications of the TRSP in the context of maintenance services are provided by \cite{Damm2021} who combine customer time windows with technician lunch breaks. They implement a genetic algorithm with the multi-objective approach of ensuring both executing priority tasks and serving customers at the beginning of their time windows. The algorithm is able to find up to 94\% on average of all Pareto-optimal solutions. \cite{Nunes2023} present a case study from a major Portuguese company providing on-site technical assistance. A cheapest insertion heuristic is compared to the currently implemented solution, showing an improved algorithmic and routing efficiency for the operational performance. \cite{Mathlouthi2021} implement a metaheuristic based on tabu search to approach the TRSP in an application for the repair and maintenance of electronic transaction equipment. They extend the problem by service time windows and an inventory of spare parts carried by each technician. Related to different objectives like travel distance or service delay, their algorithm finds the optimum with up to 50 tasks that needed to be performed whereas exact algorithms are not able to solve such larger instances in reasonable computational runtime. 

\subsection{Dynamic Multi-Period Service Routing}

\cite{Ulmer2018} present a multi-period routing problem with dynamic requests. The decision maker accepts a subset of new requests and then decides who to serve today and tomorrow, evaluating decisions via today's and tomorrow's increase in travel time. Postponed customers must be served in the next period. The goal is to maximize the number of accepted requests. With an anticipatory dynamic policy, the authors show how anticipation changes the acceptance behavior and leads to a fairer geographical service distribution. \cite{Ulmer2020} examine the tactical value of familiarity in a stochastic and dynamic TRSP, focusing on minimal travel and service times. All drivers and customer are unfamiliar with each other in the beginning, allowing them to gain familiarity through subsequent visits. Over time, drivers have familiarity with a base of customers regularly visited, reducing their required service times and increasing customer retention. By implementing an anticipatory solution methodology, the authors determine selected driver-customer pairs for which establishing short-term investments to enhance familiarity would yield the greatest benefits in the long-term. \cite{Yildiz2020} consider a multi-period vehicle routing problem where customers are offered discounts to accept alternative delivery times, aimed to gain additional flexibility for providers when delivering. With an employed demand management strategy, costs can drop significantly when customers are offered discounts in delivery fee. Moreover, it is crucial to offer individual discounts to customers instead of offering same discount to all customers. \cite{Avraham2021} introduce a multi-period technician routing problem with customer requests occurring sequentially over time. The service provider offers time slots to customers for the day and daytime of service which they can either accept or reject. Customer decide based on a discrete choice model with its parameters known to the provider. The employed solution policy improve the acceptance rate and utility for customers. \cite{Keskin2023} present a multi-period vehicle routing problem where a service provider can offer customers an incentive to request services sooner. In the context of waste collection, the authors develop a rule-based policy to decide which customers to contact and ask for an earlier pick-up of waste. Finding a balance between increased frequencies of visits with smaller demand units per visit and serving conveniently located customers on the way is challenging. Applying different strategies, for e.g., using customer characteristics or considering the current plan at the time of asking for earlier service, can reduce travel distances significantly.

\subsection{Routing Problems with Repeated Customer Visits}

\cite{Barkaoui2015} introduce a dynamic vehicle routing problem with time windows to capture the customer satisfaction level over multiple visits. An outcome of a specific visit can be a success or failure whereby the latter negatively impacts the individual customer satisfaction level. Customers need to be revisited unless a certain threshold of satisfaction is not reached. An adjusted hybrid genetic algorithm which incorporates the anticipation of future customer visits into routing decisions shows increasing numbers of satisfied customers at less traveled distances. \cite{Liu2016} present a vehicle routing problem with stochastic customer requests to examine the implications of additional distances and unloading times caused by service failures. This work relates to fields where insufficient remaining vehicle capacities represent the reasons for on-site service failures. In such cases, vehicles must return to the depot, unload their inventory and then revisit the customer which incurs additional logistic costs. The decision to continue customer visits, despite the risk of future service failures, depends on the decision-maker's willingness to take risks. The authors apply various coordination rules that require vehicles with surplus remaining capacities to assist those with limited capacities in completing services, aiming to prevent service failures. It turns out that these coordination rules are quite useful for risk-seeking decision-makers as they effectively balance a trade-off between reducing extra traveled distances caused by service failures and preventing unused vehicle capacities. \cite{SalavatiKhoshghalb2019} present an optimal restocking policy in a vehicle routing problem with stochastic customer requests. Similar to the work of \cite{Liu2016}, service failures occur if the customer demand exceeds the residual capacity of a vehicle, resulting in additional travel costs due to customer returns. The authors implement an optimal restocking policy that approximates lower bounds for the expected costs related to revisiting customers. These boundaries enable a decision-maker to choose between either preventive depot returns for replenishment in anticipation of future service failures or visiting a customer without replenishing. The latter represents a risk of longer travel distances in case of service failures as vehicles need to unload their load at the depot first before returning to the customer. By assuming customer demands to follow discrete probability distributions, the authors were the first to optimally solve larger instances with up to 60 customers and four vehicles using an exact method for the optimal restocking policy.

\section{Decision Space}\label{app: decision_space}

Several routing constraints must hold to ensure a feasible solution $x_t =(y_t,z_t)$. In the following constraint list, we only consider a single time period $t$. We define $\mathcal{K}^0_t$ as the set of customers including the depot $\{0\}$ and $h$ as the working limitation for technicians. The feasible decision space $\mathcal{X}(S_t)$ is defined as:

\begin{subequations}\label{eq: constraints}
	\begin{align}
		&\hspace{0.3cm} \sum_{w \in \mathcal{W}_t} y_{wit} \leq 1 &&\forall i \in \mathcal{K}_t &\label{eq: orienteering} && \\[1.9ex]
		&\hspace{0.3cm} \sum_{i \in \mathcal{K}^0_t} \sum_{j \in \mathcal{K}^0_t \backslash \{i\}} z_{wijt} \times {\tau_{ijt}} \leq h &&\forall w \in \mathcal{W}_t  &\label{eq: workload} && \\[1.6ex]
		&\hspace{0.25cm} \sum_{j \in \mathcal{K}^0_t} z_{w0jt} = \sum_{j \in \mathcal{K}^0_t} z_{wj0t} = 1 &&\forall w \in \mathcal{W}_t  &\label{eq: depot} && \\[1.6ex]
		&\hspace{0.3cm}\sum_{j \in \mathcal{K}^0_t \backslash \{i\}} z_{wijt} = \sum_{j \in \mathcal{K}^0_t \backslash \{i\}} z_{wjit} = y_{wit} \hspace{2.94cm} &&\forall i \in \mathcal{K}_t, w \in \mathcal{W}_t  &\label{eq: flow} && \\[1.6ex]
		&\hspace{0.3cm} \sum_{\substack{i \in \mathcal{E}}} \sum_{\substack{j \in \mathcal{E}}} z_{wijt} \leq |\mathcal{E}| - 1 &&\forall w \in \mathcal{W}_t, \mathcal{E} \subset \mathcal{K}_t &\label{eq: subtour} && \\[2ex]
		&\hspace{0.4cm} y_{wit} \hspace{3.5pt} \in \{0, 1\} &&\forall i \in \mathcal{K}_t, w \in \mathcal{W}_t  &\label{eq: domain_y} &&  \\[2.5ex]
		&\hspace{0.4cm} z_{wijt} \in \{0, 1\} &&\forall i,j \in \mathcal{K}^0_t, w \in \mathcal{W}_t &\label{eq: domain_z}  && 
	\end{align}
\end{subequations}

\noindent The initial Constraints~\ref{eq: orienteering} control that every customer is visited at most once. Constraints \ref{eq: workload} respect the maximum daily working capacity per technician. In Constraints~\ref{eq: depot}, we ensure that each technician leaves the depot and returns back to it. The flow conservation (see Constraints~\ref{eq: flow}) ensure that technicians visiting customers also leave the respective customers to continue their tours. Furthermore, we need to eliminate potential subtours to guarantee that each technician route is one single connected sequence of customer visits. Therefore, Constraints~\ref{eq: subtour} introduce subsets $\mathcal{E}$ of the complete customer set $\mathcal{K}_t$ in order to prevent the occurrence of isolated subcycles in the routing \citep{Vansteenwegen2019}. Finally, the Constraints~\ref{eq: domain_y} and \ref{eq: domain_z} define the binary variable domains. For more general background on orienteering problems, we refer to \cite{Gunawan2016}.

\section{Methodological Extension}\label{app: methodological}

In this section, we first provide the proof of our Proposition~\ref{prop: monotonicity}. Further, we provide explanation regarding the differences between the concepts of service urgency and inconvenience function. We then elaborate on the feature selection process conducted to perform our RL method. Finally, we provide the enumeration for the balancing parameter $\alpha$ applied by the \textit{SB} policy.

\subsection{Proof of Proposition~\ref{prop: monotonicity}}\label{app: proof}

Showing $V\big(S_t^{x^\prime}\big)\leq V\big(S_t^x\big)$ is by definition equivalent to showing:
    \begin{equation}\label{eq: equality_of_expected_costs}
         \mathbb{E} \Biggl[\sum_{l=t+1}^{T} \Big(C\big(S_l, X^{\pi^*} (S_l)\big)\mathrel{\Big|}S_t^{x\prime}\Big)\Biggr]\leq  \mathbb{E} \Biggl[\sum_{l=t+1}^{T} \Big(C\big(S_l, X^{\pi^*} (S_l)\big)\mathrel{\Big|}S_t^{x}\Big)\Biggr].
    \end{equation}
    Let $(S_t, x)$ and $(S_t^\prime, x^\prime)$ denote the states and decisions associated with post-decision states $S_t^x$ and $S_t^{x\prime}$.
    Given arbitrary stochastic information $\omega_{t+1}$, we construct two potential states $\bar{S}_{t+1}=\mathfrak{T}(S_t, x, \omega_{t+1})$ and $\bar{S}_{t+1}^\prime=\mathfrak{T}(S_t^\prime, x^\prime, \omega_{t+1})$. By construction, both states $\bar{S}_{t+1}, \bar{S}_{t+1}^\prime$ are identical except for the deadlines of the failed requests $\mathcal{K}^{r\omega}_{t+1}$ during transition and unassigned requests $\mathcal{K}^{xu}_t$ from the previous period. Thus, it holds again that $\delta_{t+1}\leq \delta^\prime_{t+1}$. Further, we have by construction that the transition probabilities are equal, i.e., $\mathbb{P}(\bar{S}_{t+1}^\prime\mid S_t^{x\prime})=\mathbb{P}(\bar{S}_{t+1}\mid S_t^x)$. 
    Using Equation~\eqref{eq: cost}, it is straightforward to show that $\mathbb{E}[C]$ is monotonically decreasing in $\delta_t$. 
    Thus, by applying the optimal decision in $\bar{S}_{t+1}$ to $\bar{S}_{t+1}^\prime$, we yield by the monotonicity of $\mathbb{E}[C]$:
    \begin{equation}
        \mathbb{E}\Big[C\big(\bar{S}_{t+1}^\prime, X^{\pi^*}(\bar{S}_{t+1})\big)\Big]\leq \mathbb{E}\Big[C\big(\bar{S}_{t+1}, X^{\pi^*}(\bar{S}_{t+1})\big)\Big].
    \end{equation}
    As this holds for arbitrary stochastic information and because 
    \begin{equation}
    \mathbb{E}\Big[C\big(\bar{S}_{t+1}^\prime, X^{\pi^*}(\bar{S}_{t+1}^\prime)\big)\Big]\leq \mathbb{E}\Big[C\big(\bar{S}_{t+1}^\prime, X^{\pi^*}(\bar{S}_{t+1})\big)\Big], 
    \end{equation}
    it also follows that:  
    \begin{equation}
    \begin{aligned}
    &\mathbb{E}_{\omega_{t+1} \sim \Omega}\Big[C\big(S_{t+1}, X^{\pi^*}(S_{t+1})\big) \mathrel{\Big|} S_t^{x\prime}, S_{t+1} = \mathfrak{T}(S_t^\prime, x, \omega_{t+1})\Big] \\
    \leq &\mathbb{E}_{\omega_{t+1} \sim \Omega}\Big[C\big(S_{t+1}, X^{\pi^*}(S_{t+1})\big) \mathrel{\Big|} S_t^x, S_{t+1} = \mathfrak{T}(S_t, x, \omega_{t+1})\Big].
    \end{aligned}
    \end{equation}
    
    \noindent Identity~\eqref{eq: equality_of_expected_costs} follows by repeating the same argument for every potential chain of stochastic information $(\omega_{t+1},\omega_{t+2},\dots)$ and optimal decisions $(x_{t+1}^*, x_{t+2}^*,\dots)$ applied to $S_t^x$. $\blacksquare$
\endproof

\subsection{Service Urgency and Inconvenience Function}\label{app: service_urgency}

\begin{figure}[!t]
    \centering 
    \includegraphics[width = 0.56 \textwidth]{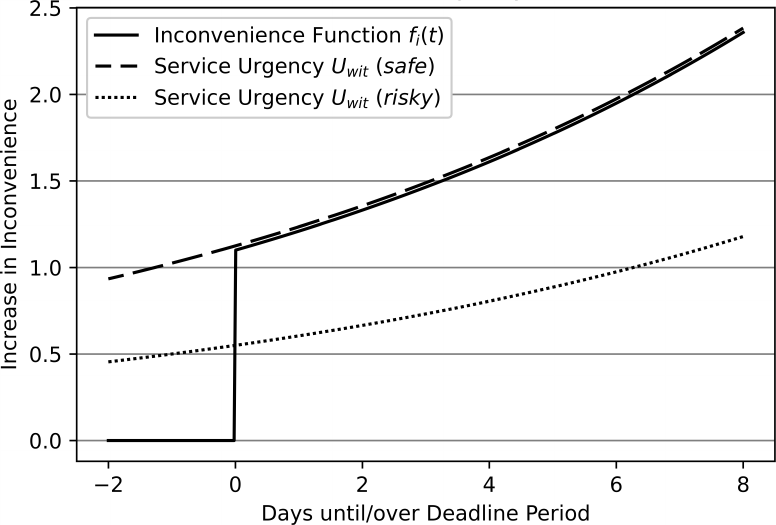}
    \caption{Impact of deadline status on increase in inconvenience with $\textbf{p = 0.5}$\centering}
    \label{fig: functions}
\end{figure}

As we have designed the service urgency $U_{wit}$, we aim to mimic the structure of the inconvenience function $f_i(t)$, as it is relevant for calculating costs (see Equation~\ref{eq: real_cost}), but adapt its definition for $t<\delta_{it}$. We visualize these differences in Figure~\ref{fig: functions}. The y-axis shows the increase in inconvenience during the transition between two periods. The value $0$ on the x-axis indicates the deadline period. The inconvenience function $f_i(t)$ depicted by the solid line shows that customers do not experience any increase in inconvenience as long as their deadlines are not reached (negative values on x-axis). Only in periods after the deadline, recognizable by positive values on the x-axis, increases in inconveniences occur for incomplete services. The expected increase in inconvenience that can be saved according to the definition of service urgency is depicted in dashed and dotted lines for \textit{safe} and \textit{risky} assignments, respectively. In the score function, this finally allows for differentiation among customers with varying levels of urgency, even if their deadlines are in the future.

\subsection{Feature Selection}\label{app: feature}

To train our parametrization function $\Lambda$, we need to consolidate the state variable $S_t$ that serves as its input. In the following, we denote our set of state features. 

\noindent\textbf{General state information}
\begin{compactitem}
    \item Time period within the horizon.
    \item Number of \textit{easy} and \textit{advanced} customer requests.
    \item Number of available \textit{regular} and \textit{expert} technicians.
\end{compactitem}

\noindent\textbf{Spatial information}
\begin{compactitem}
    \item Average distance from \textit{easy} and \textit{advanced} customer locations to the depot.
    \item Average distance between \textit{easy} and \textit{advanced} customer locations. 
\end{compactitem}

\noindent\textbf{Deadline information}
\begin{compactitem}
    \item Number of non-urgent \textit{easy} and \textit{advanced} customers.
    \item Number of overdue \textit{easy} and \textit{advanced} customers.
    \item Average number of periods deadlines have expired for overdue customers.
\end{compactitem}

\subsection{Enumeration of Balancing Parameter}\label{app: enumeration}

\begin{figure}[!t]
    \centering 
    \includegraphics[width = 0.56 \textwidth]{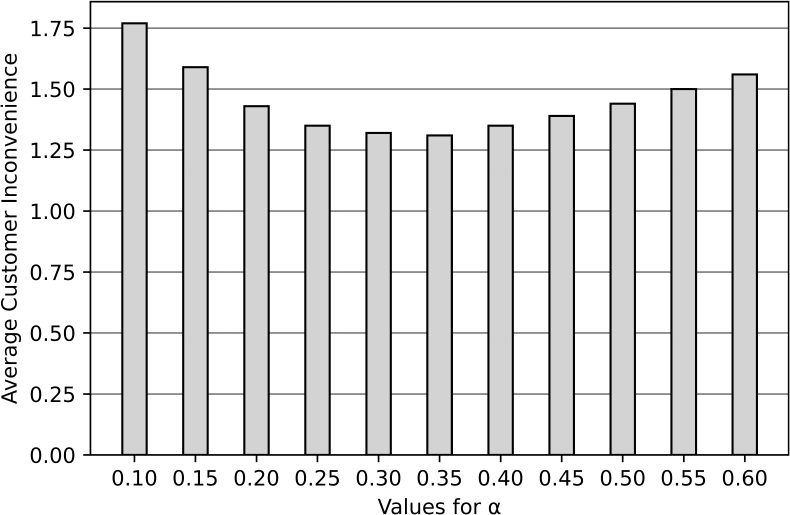}
    \caption{Grid Search for best $\boldsymbol{\alpha}$ magnitudes \centering}
    \label{fig: grid}
\end{figure}

The introduced \textit{SB} policy assumes a static and state-independent parametrization for $\alpha$. To come up with such a magnitude that promises the best objective value, we performed a grid search (see Figure~\ref{fig: grid}). For values between 0.1 and 0.6 with a step size of 0.05 (see x-axis), we iteratively ran our simulation and found the lowest average customer inconvenience (see y-axis) for values around 0.35. Ultimately, we found the best value for $\alpha$ = 0.33.

\section{Extended Results}\label{app: results}

In this section, we provide extended results from our computational evaluation. Initially, we present the accumulation of customer inconvenience over time. Then, we provide insights into real technician routes conducted by different policies. In a final section, we compare the performance of all policies across various key performance indicators as well as for different technician fleet sizes.

\subsection{Learning Exploration Rate During Training}\label{app: long_training}

In Figure~\ref{fig: rl_curve_complete}, we depict the entire learning curve for configuration (5) which we have introduced in the algorithmic augmentation details (see Table~\ref{tab: tricks}). The policy learns the standard deviation $\sigma_k$ by its own, resulting in a very slow learning process as $\sigma_k$ is only decayed gradually. After 70,000 iterations the curve flattens slowly. Yet, we still recognize marginal jumps until around 100,000 iterations. 

\subsection{Temporal Development of Inconvenience}\label{app: cummulative_inconvenience}

\begin{figure}[!t]
    \centering
    \begin{minipage}{0.49\textwidth}
        \centering
        \includegraphics[width=\textwidth]{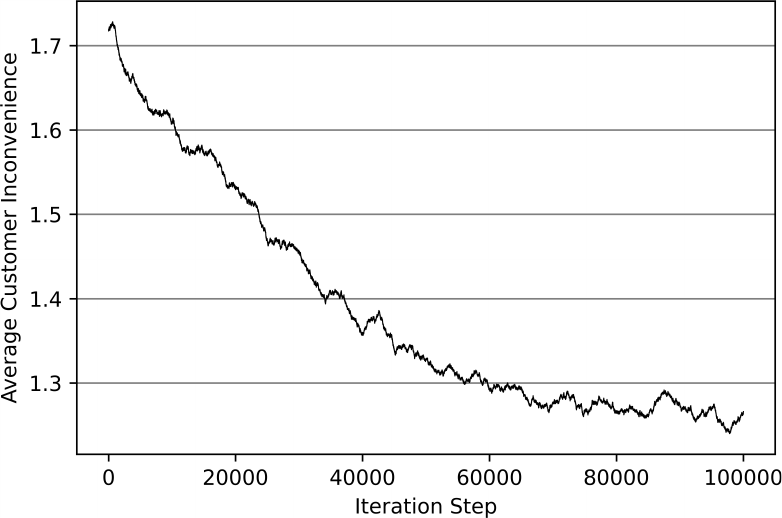}
        \caption{Learning curve for configuration (5) \centering}
        \label{fig: rl_curve_complete}
    \end{minipage}\hspace{0.2cm}
    \begin{minipage}{0.486\textwidth}
        \centering
        \includegraphics[width=\textwidth]{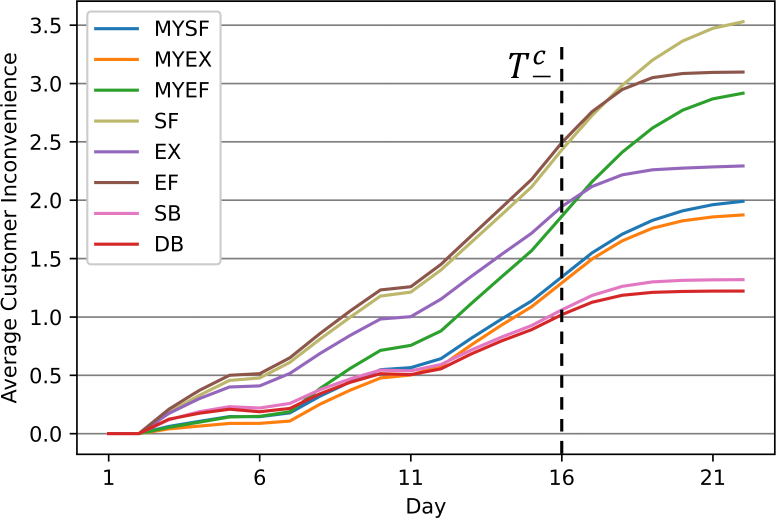}
        \caption{Increase in inconvenience over time \centering}
        \label{fig: cum_delay}
    \end{minipage}
\end{figure}

In Figure~\ref{fig: cum_delay}, the x-axis and y-axis represent the days and the average inconvenience, respectively. Time period $T^c=16$, depicted by the dashed vertical line, represents the start of the \textit{leftover} phase. From this day on, no new customer are revealed. We clearly observe the implications that come along with the triple amount of customer requests on Mondays. The kinks in the graphs around Mondays (days $1,6,11$) indicate disproportionate increases in inconvenience, particularly for benchmark policies, as the system becomes highly congested during these times. We next compare the individual policies. Policies that disregard customer deadlines (\textit{SF}, \textit{EX}, \textit{EF}) show immediate inconveniences from day $3$ on with monotonically increasing values in subsequent periods. The myopic policies \textit{MYSF}, \textit{MYEX} and \textit{MYEF} minimize the immediate inconvenience. They achieve the least inconvenience in the first $8$ periods, but then values increases substantially. This confirms the shortsightedness of policies with a focus on minimizing only the immediate inconvenience. Policies \textit{SB} and \textit{DB} show the slightest increases in inconvenience over time. In contrast to the other policies, they also show a quick convergence only a few periods after the cutoff period $T^c$. This has two reasons. First, the number of remaining customers is smaller compared to the other policies in $T^c$. However, there is another reason, the spatial distribution of the remaining customers. Especially the \textit{SF}, \textit{EX} and \textit{EF} policies mainly assign customers with respect to routing efficiency. This leaves many customers in the system who are located far from the depot (see Figure~\ref{fig: geo_delay}), impeding the fast completion of all services during the \textit{leftover} phase.

\subsection{Routing Decisions}\label{app: routing_instances}

\begin{figure}
    \centering
    \begin{subfigure}{0.49\textwidth}
    \includegraphics[width=\textwidth]{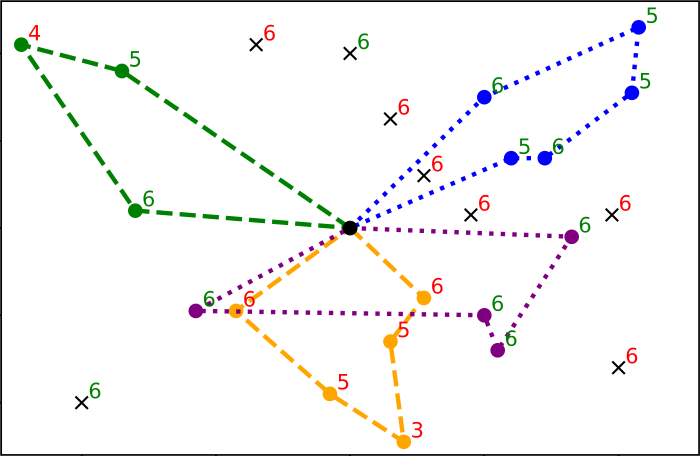}
    \caption{Routing decision by \textit{MYSF}}
    \label{fig: routing_mysf}
    \end{subfigure}
    \begin{subfigure}{0.49\textwidth}
    \includegraphics[width=\textwidth]{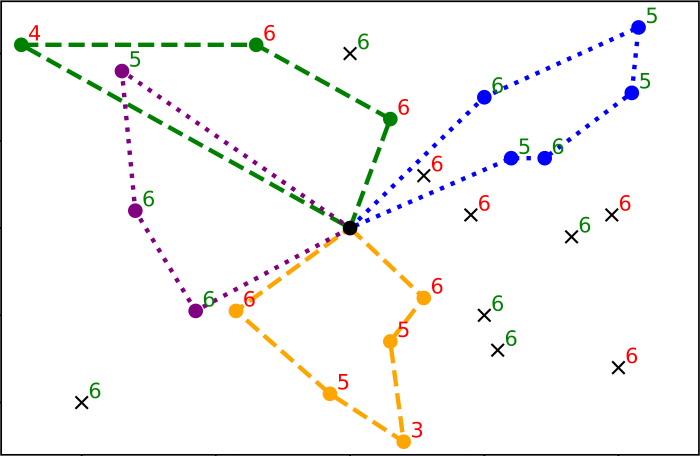}
    \caption{Routing decision by \textit{MYEX}}
    \label{fig: routing_myex}
    \end{subfigure}
    \begin{subfigure}{0.49\textwidth}
    \includegraphics[width=\textwidth]{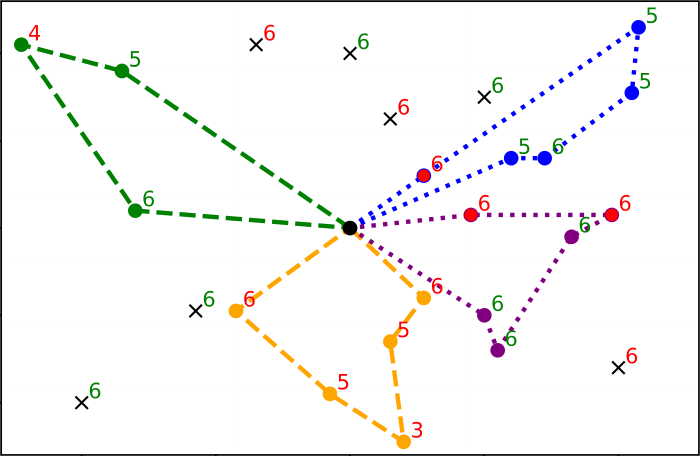}
    \caption{Routing decision by \textit{MYEF}}
    \label{fig: routing_myef}
    \end{subfigure}
    \begin{subfigure}{0.49\textwidth}
    \includegraphics[width=\textwidth]{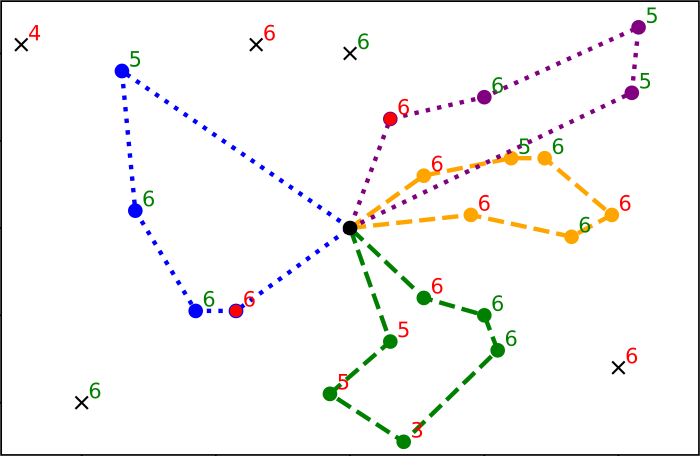}
    \caption{Routing decision by \textit{DB}}
    \label{fig: routing_db}
    \end{subfigure}
\caption{Real routing decisions for day \textbf{t=4}\centering}
\label{fig: real_instances}
\end{figure}

In Figure~\ref{fig: real_instances}, we illustrate routing and assignment decisions generated by different policies for period $t=4$ (Thursday). For the sake of clearness, we decreased the number of new requests and available technician compared to our test instances in the computational study. The black circle in the center represents the depot location, dotted (dashed) lines characterize \textit{regular} (\textit{expert}) technician tours. Numbers next to customers indicate their individual deadline periods, green and red colored numbers represent \textit{easy} and \textit{advanced} customers, respectively. Mismatched (\textit{risky}) assignments leading to potential rework can be recognized by the red circles integrated in \textit{regular} technician tours.

In the upper half, we depict decisions made by \textit{MYSF} and \textit{MYEX} policies. Both focus on minimizing the immediate customer inconvenience while preventing \textit{risky} assignments. There are two customers contributing to inconvenience if not assigned. One customer in the south with an overdue deadline in $t=3$ and one customer in the northwest with an imminent deadline in the considered period $t=4$. As these customers are assigned, both decisions induce no inconvenience in this period. However, we can clearly observe the drawbacks of such policies that ensure \textit{safe} assignments and prioritize visiting urgent customer first. Technician tours clearly overlap as \textit{advanced} customers are not assigned to \textit{regular} technicians. As working capacities for \textit{expert} technicians are limited, \textit{regular} technicians are sent to same areas to visit (isolated) remaining requests. We see this for the \textit{MYSF} policy that dispatches an \textit{expert} technician (dashed lines) south to the depot to visit five \textit{advanced} customers. Another \textit{regular} technician (dotted lines) then visits one isolated customer southwest to the depot before continuing its tour. Both, preventing mismatched assignments and serving urgent customers first comes with high travel times. Consequently, the number of unvisited customers increases (see upper half of Figure~\ref{fig: real_instances}) and states in subsequent periods congest.

The decision of the \textit{DB} policy visualized in the lower right half of Figure~\ref{fig: real_instances} shows the balance between service urgency and routing efficiency. Compared to \textit{MYEF}, the number of unvisited customers is reduced by two and the number of mismatches by one customer. We want to shortly highlight the decision-making process for one specific technician tour to explain some motivations of our policy. Considering the \textit{regular} technician tour west to the depot, four customers including one mismatch are served. An alternative route could be identical to the \textit{expert} technician route of \textit{MYEF} performed west to the depot in the same region. Indeed, one customer is served less, but the urgent customer with deadline period $t=4$ in the northwest is assigned and no inconvenience occurs. However, the \textit{DB} policy assesses it more beneficial to allow for a slight increase in inconvenience for this unvisited customer in the northwest while serving one customer more. Moreover, the assigned mismatch for customer with deadline period $t=6$ is located much closer to the depot compared to a potential mismatch for customer with deadline period $t=4$ in the northwest. This anticipatory assignment decision helps mitigate the need for resource investments in case of rework.

\subsection{Key Performance Indicators}\label{app: kpi}

Table~\ref{tab: results_app} shows the final results for several performance indicators and for different qualified fleet sizes. In the upper (lower) part, we display the results of a similar computational study where we change the skill of the workforce by reducing (increasing) the number of \textit{expert} technicians to two (four) technicians. As expected, the performances of all policies (apart from \textit{MYEX} and \textit{EX}) improve with a higher qualified workforce. With a more imbalanced fleet, these two policies loose significant performance relative to the others. As only \textit{easy}-\textit{regular} and \textit{advanced}-\textit{expert} assignments are allowed, either \textit{expert} (upper part Table~\ref{tab: results_app}) or \textit{regular} (lower part Table~\ref{tab: results_app}) technicians face a high workload of customers. Vice versa, the respective other technician group remains more often underutilized and inconvenience rates increase. Next, we want to emphasize that the \textit{DB} policy, with only three \textit{expert} technicians available, outperforms all problem-oriented benchmark policies with four available \textit{expert} technicians. Not only for the final objective value, but also for the remaining performance indicators, a \textit{DB} policy shows the most promising results across different qualified fleets. We note that the results displayed in Table~\ref{tab: results_app} are based on a trained \textit{DB} policy for three \textit{expert} technicians, demonstrating the robustness of this policy across different test instances.

\begin{table}[!th]
\centering
\begin{tabular}{lccccc}
\hline
\vspace{1pt}
\multirow{2}{*}{\begin{tabular}[c]{@{}l@{}} \\ Policy\end{tabular}} & \multicolumn{5}{c}{\begin{tabular}[c]{@{}c@{}} \\ Number of \textit{Expert} Technicians: 2 \end{tabular}} \\ \cline{2-6} \multicolumn{1}{c}{} & \begin{tabular}[c]{@{}c@{}}\\ Avg. Customer\\ Inconvenience \end{tabular} & \begin{tabular}[c]{@{}c@{}}  \\ Avg. Customer\\ Delay (Days)\end{tabular} & \begin{tabular}[c]{@{}c@{}}  \\ Number of \\ Returning Visits \end{tabular} & \begin{tabular}[c]{@{}c@{}}  \\ \textit{Leftover} \\ Phase (Days) \end{tabular} & \begin{tabular}[c]{@{}c@{}}  \\ Invested Resources \\ (Technician-Days) \end{tabular}\\ \hline  \\
$\textit{MYSF}$ & 6.59 & 3.57 & \hspace{0.35cm} 0.00   & \hspace{0.0cm} 17.77 & \hspace{0.0cm} 110.14 \\
$\textit{MYEX}$ & 6.01 & 3.32 & \hspace{0.35cm} 0.00   & \hspace{0.0cm} 17.25 & \hspace{0.0cm} 110.18 \\
$\textit{MYEF}$ & 4.21 & 3.08 & \hspace{0.00cm} 131.41 & \hspace{0.0cm} 10.31 & \hspace{0.0cm} 127.34 \\
$\textit{SF}$   & 8.78 & 3.43 & \hspace{0.35cm} 0.00   & \hspace{0.0cm} 15.86 & \hspace{0.2cm} 98.85  \\
$\textit{EX}$   & 5.06 & 2.31 & \hspace{0.35cm} 0.00   & \hspace{0.0cm} 12.56 & \hspace{0.2cm} 98.55  \\
$\textit{EF}$   & 4.08 & 2.20 & \hspace{0.00cm} 136.75 & \hspace{0.2cm} 7.40  & \hspace{0.0cm} 109.88 \\
$\textit{SB}$   & 2.45 & 1.79 & \hspace{0.20cm} 80.57  & \hspace{0.2cm} 8.03  & \hspace{0.0cm} 114.75 \\
$\textit{DB}$   & 2.31 & 1.68 & \hspace{0.20cm} 74.61  & \hspace{0.2cm} 7.76  & \hspace{0.0cm} 113.10 \\ \hline
& \multicolumn{5}{c}{\begin{tabular}[c]{@{}c@{}} \\ Number of \textit{Expert} Technicians: 3 \end{tabular}} \\ \hline  \\
$\textit{MYSF}$ & 1.99 & 1.57 & \hspace{0.20cm} 0.00   & \hspace{0.0cm} 8.07  & \hspace{0.0cm} 108.57 \\
$\textit{MYEX}$ & 1.86 & 1.50 & \hspace{0.20cm} 0.00   & \hspace{0.0cm} 7.33  & \hspace{0.0cm} 109.08 \\
$\textit{MYEF}$ & 2.92 & 2.26 & \hspace{0.00cm} 86.83  & \hspace{0.0cm} 8.30  & \hspace{0.0cm} 118.17 \\
$\textit{SF}$   & 3.54 & 1.85 & \hspace{0.20cm} 0.00   & \hspace{0.0cm} 8.19  & \hspace{0.2cm} 96.60  \\
$\textit{EX}$   & 2.27 & 1.35 & \hspace{0.20cm} 0.00   & \hspace{0.0cm} 5.31  & \hspace{0.2cm} 97.04  \\
$\textit{EF}$   & 3.08 & 1.74 & \hspace{0.00cm} 89.90  & \hspace{0.0cm} 5.89  & \hspace{0.0cm} 103.05 \\
$\textit{SB}$   & 1.31 & 1.06 & \hspace{0.00cm} 21.73  & \hspace{0.0cm} 5.62  & \hspace{0.0cm} 103.45 \\
$\textit{DB}$   & 1.21 & 0.97 & \hspace{0.00cm} 21.41  & \hspace{0.0cm} 5.33  & \hspace{0.0cm} 101.56 \\ \hline 
& \multicolumn{5}{c}{\begin{tabular}[c]{@{}c@{}} \\ Number of \textit{Expert} Technicians: 4 \end{tabular}} \\ \hline  \\
$\textit{MYSF}$ & 1.41 & 1.18 & \hspace{0.20cm} 0.00   & \hspace{0.2cm} 6.05  & \hspace{0.0cm} 105.62 \\
$\textit{MYEX}$ & 5.10 & 2.96 & \hspace{0.20cm} 0.00   & \hspace{0.0cm} 16.05 & \hspace{0.0cm} 109.75 \\
$\textit{MYEF}$ & 2.02 & 1.63 & \hspace{0.00cm} 53.95  & \hspace{0.2cm} 6.87  & \hspace{0.0cm} 111.20 \\
$\textit{SF}$   & 2.06 & 1.22 & \hspace{0.20cm} 0.00   & \hspace{0.2cm} 4.87  & \hspace{0.2cm} 94.33  \\
$\textit{EX}$   & 4.52 & 2.15 & \hspace{0.20cm} 0.00   & \hspace{0.0cm} 11.59 & \hspace{0.2cm} 98.31  \\
$\textit{EF}$   & 2.36 & 1.38 & \hspace{0.00cm} 54.09  & \hspace{0.2cm} 4.81  & \hspace{0.2cm} 98.11  \\
$\textit{SB}$   & 0.91 & 0.77 & \hspace{0.20cm} 7.03   & \hspace{0.2cm} 4.54  & \hspace{0.2cm} 98.58  \\
$\textit{DB}$   & 0.88 & 0.73 & \hspace{0.20cm} 8.11   & \hspace{0.2cm} 4.19  & \hspace{0.2cm} 96.89  \\ \hline\hline
\end{tabular}
\vspace{0.2cm}
\caption{Policy performances for different qualified working fleets}
\label{tab: results_app}
\end{table}

\end{appendices}

\newpage
\bibliographystyle{apalike}  
\bibliography{main}  

\end{document}